# Obstacles and Opportunities for Learning from Demonstration in Practical Industrial Assembly: A Systematic Literature Review

**Victor Hernandez Moreno[1,@], Steffen Jansing[2], Mikhail Polikarpov[2], Marc G. Carmichael[3], Jochen Deuse[1,2]**

[1] Centre for Advanced Manufacturing, Faculty of Engineering and Information Technology, University of Technology Sydney, 81 Broadway, Ultimo 2007, Australia

[2] Institute of Production Systems, TU Dortmund University, Leonhard-Euler-Str. 5, Dortmund 44227, Germany

[3] Robotics Institute, Faculty of Engineering and Information Technology, University of Technology Sydney, 81 Broadway, Ultimo 2007, Australia

[@] Victor.HernandezMoreno@student.uts.edu.au

## Abstract

Learning from demonstration is one of the most promising methods to counteract the challenging long-term trends in repetitive industrial assembly. It offers not only a programming technique that is accessible to workers on the shop floor, reducing the need for robot experts and the associated costs but also a possible solution to the observable shift from mass-production to mass-customisation through flexible and generalising systems. Since the emergence of the learning from demonstration idea in the 1980s, its methodologies, capabilities, and achievements have constantly evolved. However, despite reports of continued progress in academic publications, the concept has not yet robustly emerged across the assembly industry. In light of its great potential, this paper presents the findings from a systematic literature review following the updated Preferred Reporting Items for Systematic Reviews (PRISMA) guidelines. It aims to provide an overview of the state-of-the-art learning from demonstration solutions developed for assembly-related tasks and offer a critical discussion of remaining obstacles in order to drive its progression towards meaningful deployments. The analysis includes a total of 61 papers over the period of 2013-2023 sourced from Scopus and Web of Science databases. Findings indicate that learning from demonstration has attained a significant level of maturity within the research environment, as evidenced by thorough experimental achievements, proving its great promise for industrial assembly applications. However, critical obstacles exist in the area of proven practicability, task complexity and diversity, generalisation, performance evaluation and integration concepts that require attention to promote its widespread adoption and create a seamless transition into industrial practices.

## Keywords

Robotic Assembly, Repetitive Industrial Assembly, Robotic Application, Literature Review

## 1. Introduction

Assembly is one of the most important processes in the manufacturing industry. It converges the upstream design, engineering, manufacturing, and logistics processes to create a product that fulfils the designated function. Within the production process, the time allocated for the assembly accounts for a significant share ranging between 15% to 70% (Lotter & Wiendahl, 2012). In detail, the share in mechanical engineering varies between 20 and 45 %, depending on the product's complexity, while in vehicle construction, the assembly is estimated around 30 to 50 % of the total production time, contingent on vertical range of manufacture. The highest proportion, 40 to 70 %, is found in electrical engineering and precision mechanics. However, repetitive industrial assembly faces emerging factors that threaten to limit its growth. While the assembly process remains dominantly performed manually (Petzoldt et al., 2023), the availability of the workforce is expected to be diminished in the near future due to demographic changes (Langhoff, 2009; Thun et al., 2007). Simultaneously, a shift from mass-production to mass-customisation is observable (Christensen et al., 2021; Pedersen et al., 2016), which limits the applicability of customised assembly systems and products tend to generally shrink in dimensions, making manual assembly more challenging or even impossible in certain industries (Nof, 2009).

In the new era of collaborative robots, the robotics industry has recognised the necessity of bringing robots closer to human operators by ensuring the operators' safety and simplifying the robots' operability. While traditional programming techniques, including lead-through and offline programming, still prevail in industrial settings (Villani et al., 2018), great progress has been made in programming interfaces enabling intuitive and more natural means to transfer intentions to the robotic system. Besides practical teach pendants and walk-through technology, advanced research efforts have been reported on vision-based methods, vocal commanding, as well as augmented and virtual reality (Ajaykumar et al., 2022; Villani et al., 2018), extending to a few commercialised assistive tools such as the TracePen (Wandelbots, 2023). Nevertheless, the factories of tomorrow will require more flexible and autonomous solutions to cope with the above-mentioned challenging trends.

One potential solution is depicted by the concept of Learning from Demonstration (LfD). LfD and conceptually comparable approaches such as learning from observation, programming from demonstration, imitation learning, or apprenticeship learning, endow robots to learn new tasks from human demonstration (Argall et al., 2009). In detail, it refers to the competence of robots to autonomously perform new tasks by observing a human's performance, learning the generalised skill, and reproducing it afterwards, even under slightly different circumstances. Thus, LfD offers an intuitive way for non-robot experts and non-technical personnel to program robots and promises to deal autonomously with dynamic environments and product variants. Despite its great potential, developed solutions to the concept proposed in several academic publications and a few commercial products, such as MIRAI (Micropsi Industries, 2023), are still not yet widely deployed in the assembly industry.

The aim of this paper is to investigate why LfD solutions have not yet gained a significant foothold in the repetitive assembly industry. This work is intended to provide interested parties from industry and research insights into the state-of-the-art in order to deliberately build jointly on it and actively shape the path of LfD solutions to industrial deployment. We assert that to increase the potential for acceptance and willingness to embed such solutions into the industrial environment, robotic systems will have to mimic the capabilities of an untrained human worker in terms of learning and execution skills. Thus, this paper seeks to answer the following research questions:

- RQ1: What are the prevalent LfD approaches in academia for tasks related to assembly?
- RQ2: What are the main research areas and problem domains that have been investigated?
- RQ3: How do state-of-the-art LfD approaches align with the training techniques and learning behaviour of human operators in the industry?
- RQ4: What are the primary obstacles that hinder the practical implementation of LfD solutions in the traditional repetitive assembly industry?

Our review can be distinguished from related review papers listed in Table 1 as follows: (Sosa-Ceron et al., 2022) reviews LfD regarding particularities of human-robot collaboration. (Z. Liu et al., 2022) discusses an extended review of robot learning in which LfD represents one possibility. (C. Chen et al., 2022) summarises efforts towards robot grinding/polishing, while (Lobbezoo et al., 2021) focuses on pick-and-place operations through Reinforcement Learning. (Billard et al., 2016; Fang et al., 2019;

Ravichandar et al., 2020; Vakanski & Janabi-Sharifi, 2017) provide a comprehensive review of general LfD aspects with a minor focus on applications. Considering robotic assembly, (Zhu & Hu, 2018) outlines similarly general features of methods for this niche without critically assessing its potential in realistic industrial scenarios. (Fang et al., 2019; Hussein et al., 2017) emphasise peculiarities of imitation learning. (Saveriano et al., 2021) represents an exemplary review of one selective LfD method in which assembly tasks are discussed as a subcategory of application scenarios. By focusing on the analysis of experimental implementation and comparison of the state-of-the-art LfD methods with industry requirements in the area of assembly-related tasks, our paper provides unique insights and a valuable contribution to the field's required progression.

*Table 1: Related review and survey publications on learning from demonstration*

| Nr. | Reference | Title | Year |
|---|---|---|---|
| **1** | (Sosa-Ceron et al., 2022) | Learning from demonstrations in human-robot collaborative scenarios: A survey | 2022 |
| **2** | (Z. Liu et al., 2022) | Robot learning towards smart robotic manufacturing: A review | 2022 |
| **3** | (C. Chen et al., 2022) | Intelligent learning model-based skill learning and strategy optimisation in robot grinding and polishing | 2022 |
| **4** | (Saveriano et al., 2021) | Dynamic movement primitives in robotics: A tutorial survey | 2021 |
| **5** | (Lobbezoo et al., 2021) | Reinforcement learning for pick and place operations in robotics: A survey | 2021 |
| **6** | (Ravichandar et al., 2020) | Recent advances in robot learning from demonstration | 2020 |
| **7** | (Fang et al., 2019) | Survey of imitation learning for robotic manipulation | 2019 |
| **8** | (Calinon, 2018) | Learning from Demonstration (Programming by Demonstration) | 2018 |
| **9** | (Zhu & Hu, 2018) | Robot learning from demonstration in robotic assembly: A survey | 2018 |
| **10** | (Hussein et al., 2017) | Imitation learning: A survey of learning methods | 2017 |
| **11** | (Vakanski & Janabi-Sharifi, 2017) | Robot learning by visual observation | 2017 |
| **12** | (Billard et al., 2016) | Handbook of Robotics – Chapter 74: Learning from Humans | 2016 |

The paper is addressed to knowledgeable readers familiar with the basic concept of Learning from Demonstration and industrial assembly. For those who consider themselves newcomers to these areas or want to refresh their knowledge, we recommend the sources (Billard et al., 2016; Ravichandar et al., 2020) and (Nof et al., 1997), respectively.

The remainder of the paper is structured as follows: Section 2 describes the methodology of how the systematic literature review was conducted. Section 3 summarises the findings acquired by reviewing the academic literature. Their critical assessment regarding transferability to an industrial environment is discussed in Section 4 and a conclusion of the review is drawn in Section 5.

## 2. Methodology

To answer the specified research questions, we chose to apply an evidence-based approach in the form of a systematic literature review, known to produce reliable, reproducible, and transparent research outcomes with minimised bias and errors (Denyer & Tranfield, 2009; Page et al., 2021; Snyder, 2019). The present systematic literature review was conducted based on the updated Preferred Reporting Items for Systematic Reviews (PRISMA 2020) guideline (Page et al., 2021). The following subsections report on the established protocol describing the chosen information sources, search strategy, eligibility criteria, and selection and data collection process.

## 2.1 Information Sources and Search Strategy

The systematic literature review was built upon the interrogation of the well-recognised databases Scopus and Web of Science (WoS) (Score, 2009). Web of Science has been searched using the *Core Collection* and the *Exact Search* option.

The search strategy was based on conceptual boundaries and the selection of appropriate filters. Seeking representative records for the state-of-the-art in assembly applications of robot Learning from Demonstration, both aspects were included in the search string with their synonymously used terminology (see Table 2). Assembly-related terms were established through a preliminary screening of abstracts that included the term “assembly” combined with the assembly taxonomy of (Nof et al., 1997). For simplicity, the consortium of conceptually comparable approaches to Learning from Demonstration will be abbreviated with the acronym LfD* throughout the remaining paper. The consecutive selection of relevant reports was achieved through the use of targeted automatic filter mechanisms as outlined in Table 2 and manual screening based on the eligibility criteria listed in Section 2.2. The final database access was performed on the 31st of March, 2023.

*Table 2: Information Sources and Search Strategy*

| Database | Search String (LfD* AND assembly-related) and Filter Parameters |
|---|---|
| **Scopus** | TITLE-ABS-KEY( ( ( learning OR programming OR teaching ) PRE/2 ( demonstration OR observation ) ) OR ( ( imitation OR apprenticeship ) PRE/2 learning ) AND robot* ) **AND** TITLE-ABS-KEY ( assembl* OR ( peg W/2 ( hole OR insertion ) ) OR interlocking OR ( pick W/1 place ) OR rivet* OR wiring OR fastener OR jamming OR glue OR gluing OR ( reach* W/2 grasp* ) OR weld* OR stacking OR screw* OR retainer OR ( ( press OR snap ) W/1 fit ) OR adhesiv* OR crimp* ) |
| | Publication Year: 2013 – 2023<br>Subject Area: Computer Science, Engineering, Mathematics<br>Language: English<br>Document Type Exclusion: Conference Review, Editorial |
| **Web of Science** | TS=( ( ( ( learning OR programming OR teaching ) NEAR/2 ( demonstration OR observation ) ) OR ( ( imitation OR apprenticeship ) NEAR/2 learning ) AND robot* ) **AND** ( assembl* OR ( peg NEAR/2 ( hole OR insertion ) ) OR interlocking OR ( pick NEAR/1 place ) OR rivet* OR wiring OR fastener OR jamming OR glue OR gluing OR ( reach* NEAR/2 grasp* ) OR weld* OR stacking OR screw* OR retainer OR ( ( press OR snap ) NEAR/2 fit ) OR adhesiv* OR crimp* ) ) |
| | Publication Year: 2013 – 2023<br>Subject Area: Robotics, Computer Science, Automation Control Systems, Engineering, Mathematics<br>Language: English |

## 2.2 Eligibility Criteria

In light of the industry challenges sought to be tackled with LfD*, publications are considered eligible that explore solutions in which a new assembly-related task is demonstrated by a human operator and successfully reproduced by a robotic system independently. Relatable experimental results are essential for assessing the state-of-the-art with regards to end-to-end solutions of assembly operations. Therefore, we intentionally include all research efforts within the scope of the scenario that provide human-performed demonstrations and reproduction by physical robots. While the existence of physical experiments is required, a practical implementation in an industrial environment is not necessary.

In order to establish coverage of high-quality publications, reputation criteria in the form of a grouping approach inspired by (Crossan & Apaydin, 2010) was applied. The first group of eligible publications includes all records published in 2022 despite citation count to acknowledge and emphasise the most recent efforts in the field of interest. Records published between 2013 and 2021 are recognised as eligible should the threshold of at least three citations on average per year be reached.

Consequently, the following exclusion criteria (EX1-6) were defined, which supplement the filter settings of Table 2:

- EX1: The study was published between 2013 – 2021, but the annual average citation count is below the threshold at the time of final database access (*citation requirement*).
- EX2: The terms LfD* and assembly-related are used in an unrelated context.
- EX3: The full text of the study is not available.
- EX4: The study is a review or survey.
- EX5: The study deals with human-robot collaborative applications in which the task is reproduced jointly.
- EX6: The study has not evaluated the proposed method with regards to physical robot execution.
- EX7: The study has not evaluated the proposed method with regards to human-performed demonstration.

### 2.3 Selection and Data Collection Process

The systematic literature review was performed following the three consecutive phases of the PRISMA 2020 (Page et al., 2021) guideline, namely identification, screening, and data collection (see Figure 1). Within the initial identification phase, the interrogation of the selected databases Scopus and Web of Science, resulted in n = 330 and n = 265 identified records, respectively. The automated filtering system of both databases, following the specified settings in Table 2 removed a total of n = 136 records. By using software and manual comparison of authors, title, and abstract, n = 190 records were identified as duplicates and merged. Furthermore, the average annual citation value was calculated for each record, which marked a total of n = 156 publications as ineligible (EX1). Within the consecutive screening phase, the preliminary revision of the title and abstract revealed n = 12 records being unrelated to the field of interest (EX2), while the full text of all records was accessible (EX3). The full reports were assessed for eligibility following the exclusion criteria EX4 – EX7. In this context, n = 4 reports were identified as reviews or surveys (EX4), n = 17 reports were solely discussing human-robot collaboration (EX5), and n = 15 + 4 reports did not evaluate the proposed method in an end-to-end solution with human and robot performing the task physically (EX6 + EX7). Consequently, a total of n = 61 studies were verified as appropriate to the present study and accordingly used for the data collection process. Appendix A provides a corresponding table of all included studies of the conducted literature review.

The included studies were reviewed for the following data: Learning from Demonstration methods, applications, and experimentally evaluated capabilities. Particular attention is paid to the presented experimental evaluation and results. The first author primarily conducted the selection and data collection processes. However, uncertainties during the processes and excluded reports were discussed and agreed to in the plenary of all authors.

## 3. Results

This Section presents the findings of the systematic literature review of academic research studies and seeks answers to the research questions RQ1+2. It aims to provide a concise understanding of the current documented interests, developments, and achievements in research, emphasising LfD* in physical assembly-related applications. Appendix A provides a comprehensive outline of all reports and their discussed characteristics. The following is structured according to meaningful quantitative and in-depth qualitative aspects.

## 3.1 Quantitative Results

The quantitative analysis of the identified collection of literature emphasises the conducted experimental evaluations and underlying LfD* methods. It outlines a comprehensive statistical overview of utilised LfD* methods, investigated application scenarios, performed assembly skills, achieved generalisability capabilities, and reported performances.

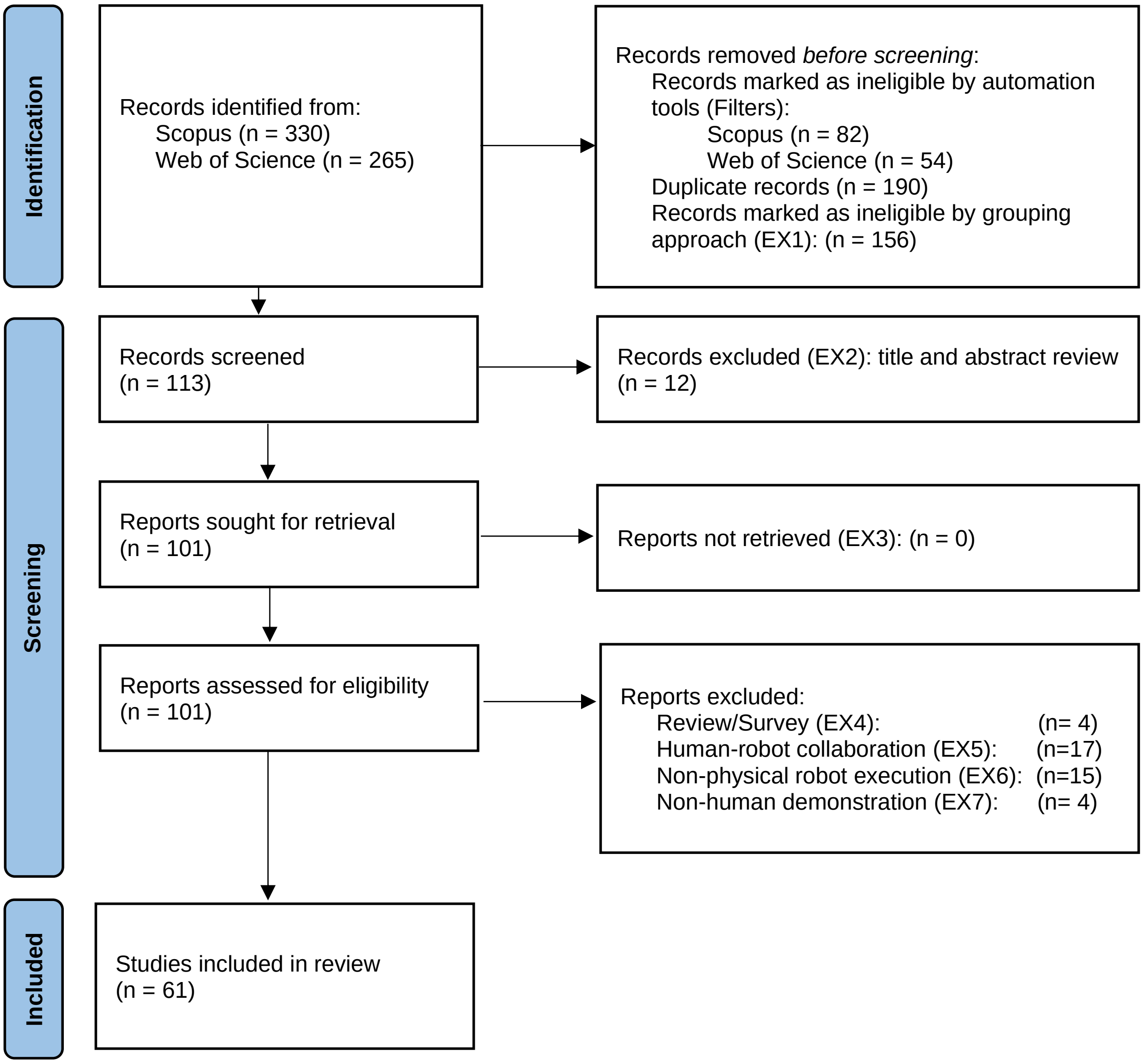

*Figure 1: PRISMA 2020 flow diagram* (Page et al., 2021)

### 3.1.1 Applied Learning from Demonstration Methods

Learning from Demonstration implementations are commonly organised into three major phases: human demonstration, model learning, and task application (Vakanski & Janabi-Sharifi, 2017). The following outlines statistical results regarding applied demonstration and learning methods according to the categorisation suggested by (Ravichandar et al., 2020).

For demonstrating potential tasks through human motions, (Ravichandar et al., 2020) proposes the categories: *kinesthetic teaching* (human moves the passive robot directly), *teleoperation* (human moves actuated robot remotely) and *passive observation* (captures robot-independent task demonstration). Figure 2(a) illustrates the resulting statistical comparison of the methods applied for human demonstration. As can be seen, *kinesthetic teaching* and *passive observation* prevail with 41% and 30% as preferred methods to teach assembly-related tasks. *Teleoperation* was exclusively selected only in 13 out of 56 cases. Five studies provided two demonstration methods jointly, either for initial skill acquisition and testing (Ji et al., 2021; Savarimuthu et al., 2018), consecutive skill correction (Meszaros et al., 2022), or teaching distinctive task aspects (Gu et al., 2018). (Abu-Dakka et al., 2015) used *teleoperation* and *kinesthetic teaching* to meet the requirements of different robotic platforms. In general, *passive observation* was achieved in various ways. The most common approach is to use camera streams or images of the recorded human demonstration (e.g. (Duque et al., 2019)). However, other studies used customised demonstration tools (Pellois & Brüls, 2022), sensor-augmented objects (Ti et al., 2022), mock-up objects with distinctive properties (e.g. a lighter object than what the robot handles (Wan et al., 2017)), or tangible instruction "blocks" (Sefidgar et al., 2017). Similarly, *teleoperation* was realised through the robot's teach pendant (Su et al., 2022), commercial tools (Davchev et al., 2022), or by mimicking the human manipulation path with identical objects in real-time (Savarimuthu et al., 2018).

In addition to the method of demonstration, the number of demonstrations required is considered an important indicator of applicability. Within the 61 analysed studies, a tendency is noticeable towards requiring two to ten demonstrations which was considered in 26 cases. On the contrary, 15 studies built on a single demonstration and eight experimental evaluations required more than ten instructions. The remaining twelve studies have not quantified (using paraphrases like "multiple", "set", "few", or "several") or not specified at all the required number of demonstrations.

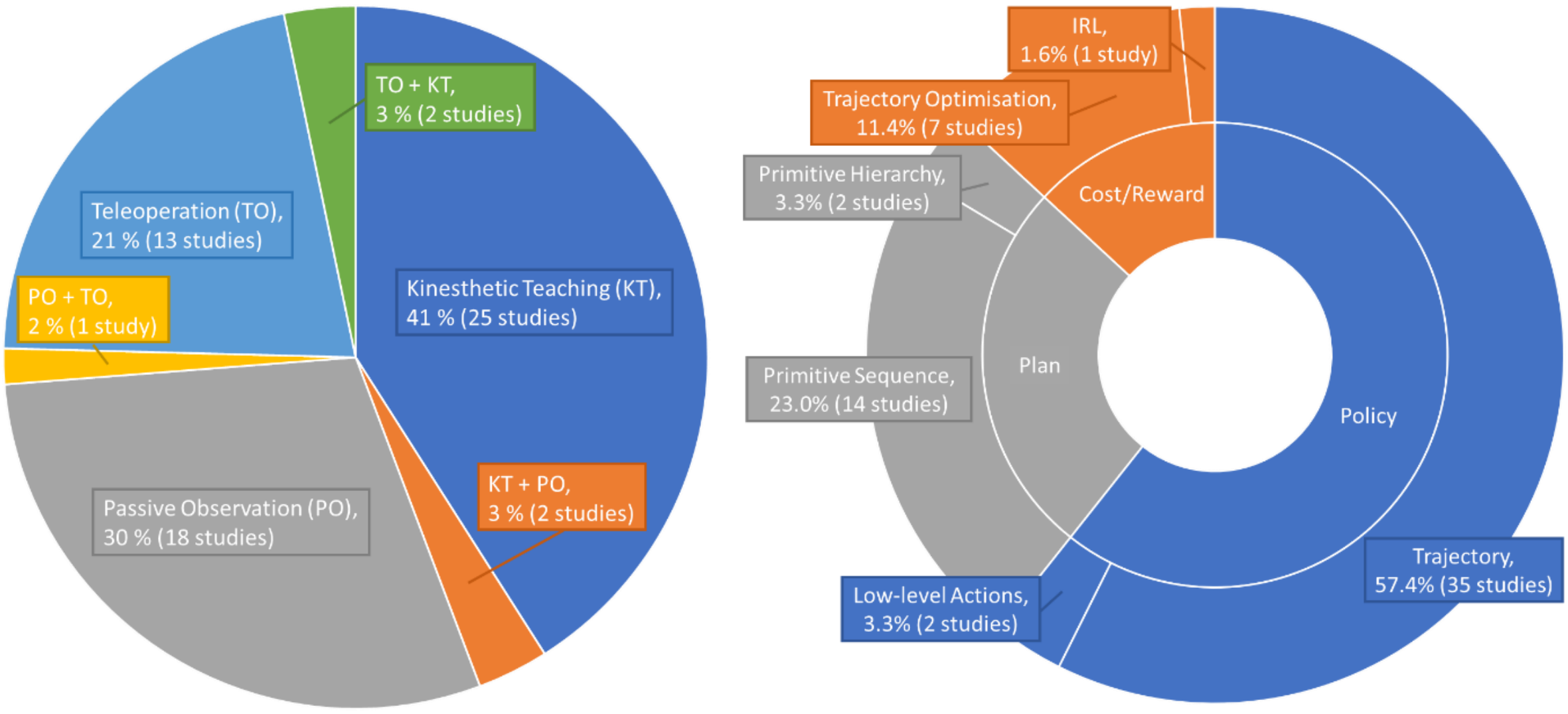

(a) Demonstration methods (five studies used a combination of two demonstration methods)

(b) Learning methods based on the learning outcome

*Figure 2: Classification of applied demonstration and learning methods after (Ravichandar et al., 2020) (see Appendix A for details)*

The categorisation of the learning methods was conducted based on the learning outcome, which can be a *policy* in the form of trajectories or low-level actions, *plan* consisting of primitive sequences or primitive hierarchies, or *cost/reward* for trajectory optimisation or inverse Reinforcement Learning (IRL) (Ravichandar et al., 2020). As can be seen in Figure 2(b), learning methods based on trajectory policies as learning outcome prevail in assembly-related LfD* research, endorsed by 35 out of 61 studies that used it as their selected approach. The second most prominent method is task representation in the form of plans based on primitive sequences reported in 23.0% of the studies.

The least reported category of applied learning methods is based on cost/reward-driven outcomes with a joint share of 13.1%. Among the most prominent techniques are so-called Dynamic Movement Primitives and Reinforcement Learning. These techniques were explored in a total of 19 and 14 studies, respectively, and applied both independently and in combination with other techniques. The analysis of learning methods in terms of preferred demonstration methods shows furthermore that 54.3% of the studies used kinesthetic teaching when targeting trajectory policy outcomes, while 64.3% preferred passive observation for outcomes in the form of primitive sequences (see Appendix A).

### 3.1.2 Application Scenarios

In the context of application scenarios, the experimental reports are analysed regarding their practicability for real-world scenarios and to which extent research approaches respond to actual assembly scenarios in the industrial sector. Therefore, three categories of practicability are defined. Studies that evaluate their LfD* method in a practical industrial scenario assembling realistic objects are categorised as "practical". The second level of practicability considers handling "related" objects. This includes objects that are only handled in a subsidiary manner in the industry, objects interesting for specific industry sectors but not practically applied, or benchmark models mimicking industrial challenges. All experiments utilising objects not meeting the above categories are considered "unrelated".

The collection of identified studies features six practical assembly scenarios with realistic objects that mainly target the electronics industry (see Figure 3). (Hu et al., 2021, 2022) investigated the PCB assembly on the bottom case of a cursor mouse, requiring fitting two locating pins and three resilient fasteners (see Figure 3(a)). Similarly, (Haage et al., 2017) investigated PCB assembly based on visual passive observation. (Yan Wang et al., 2021a, 2021b) conducted experiments on a condenser assembly task of a circuit breaker (see Figure 3(c)) that required an L-shaped insertion motion. More complex task sequences in an industrial scenario were emphasised by (Ji et al., 2021), who evaluated their LfD* solution on the assembly of power breakers and set-top boxes (see Figure 3(d)). Precision insertion and gluing capabilities for joining micro sleeve-cavities and coil-cylinders with $10\,\mu m$ clearance fit were explored by (Qin et al., 2019). Finally, (Yue Wang et al., 2018) investigated the assembly of a switch through passive observation including placing, screwing, and pushing motions.

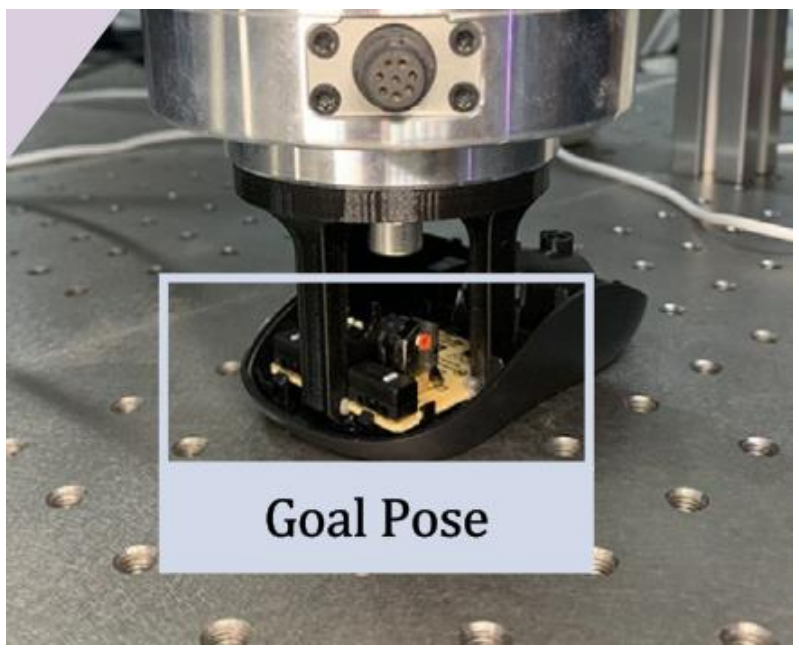

(a) PCB assembly of cursor mouse (Hu et al., 2021, 2022)

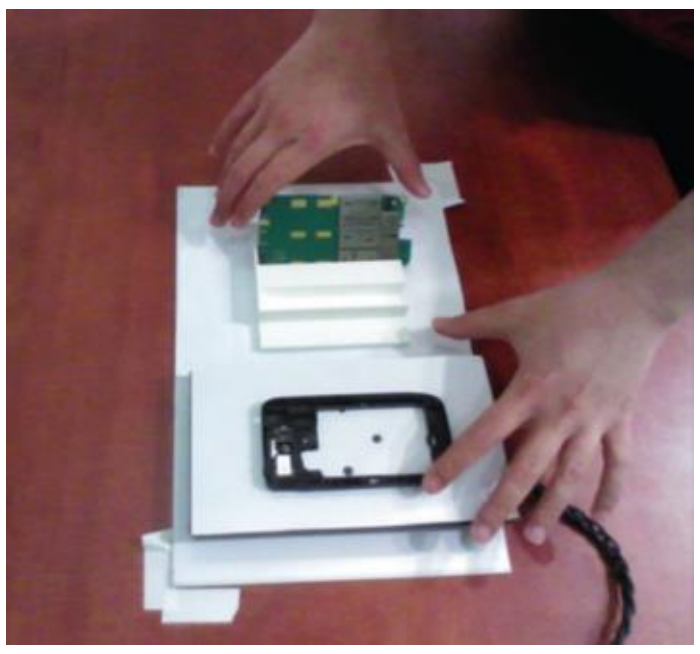

(b) PCB assembly (Haage et al., 2017)

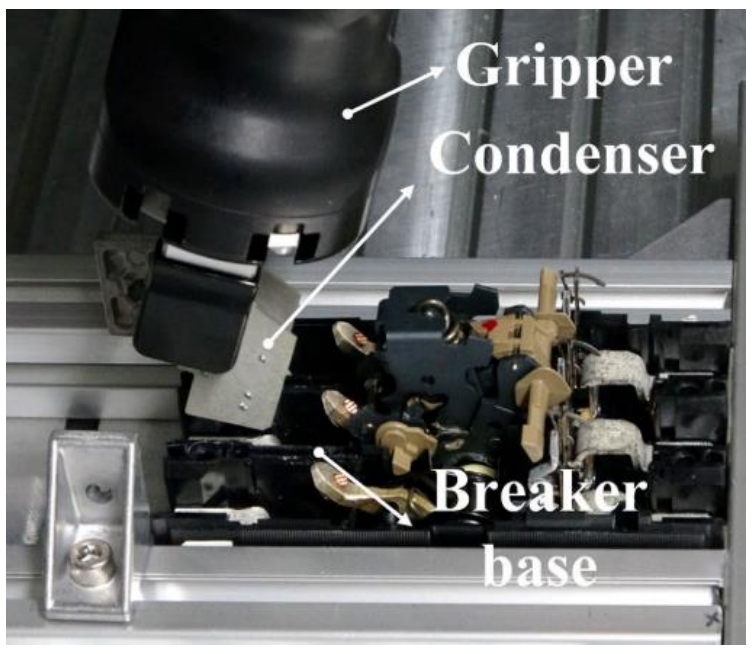

(c) condenser assembly (Yan Wang et al., 2021a, 2021b)

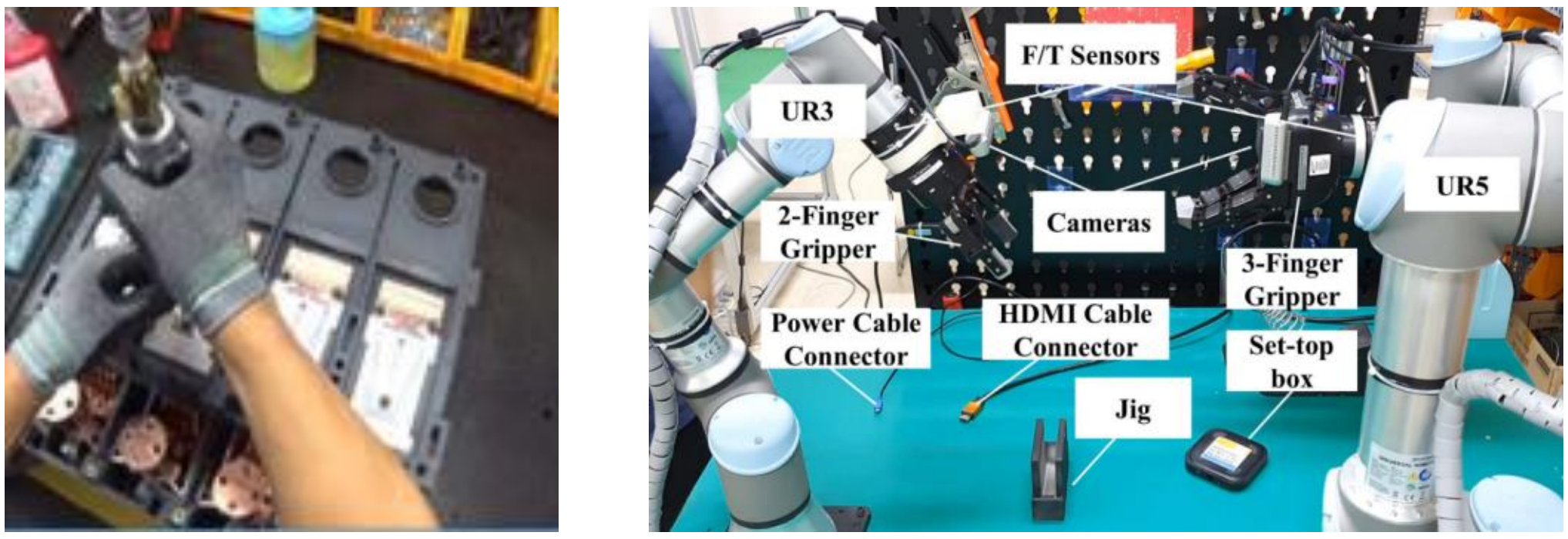

(d) power breaker (left) and set-top box (right) assembly (Ji et al., 2021)

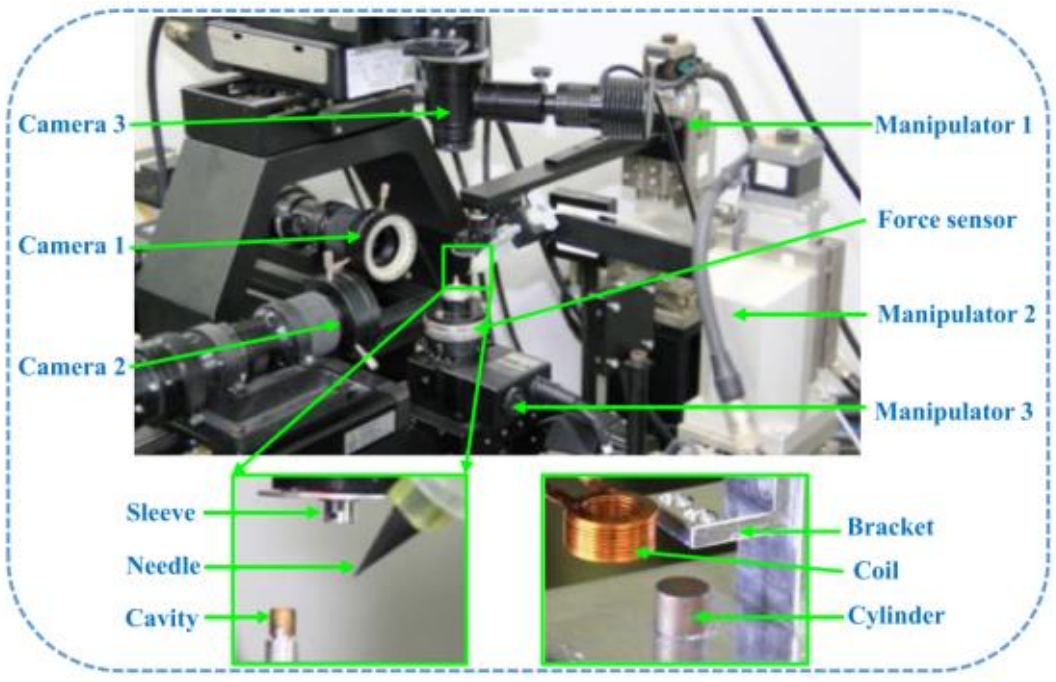

(e) sleeve-cavity and coil-cylinder assembly (Qin et al., 2019)

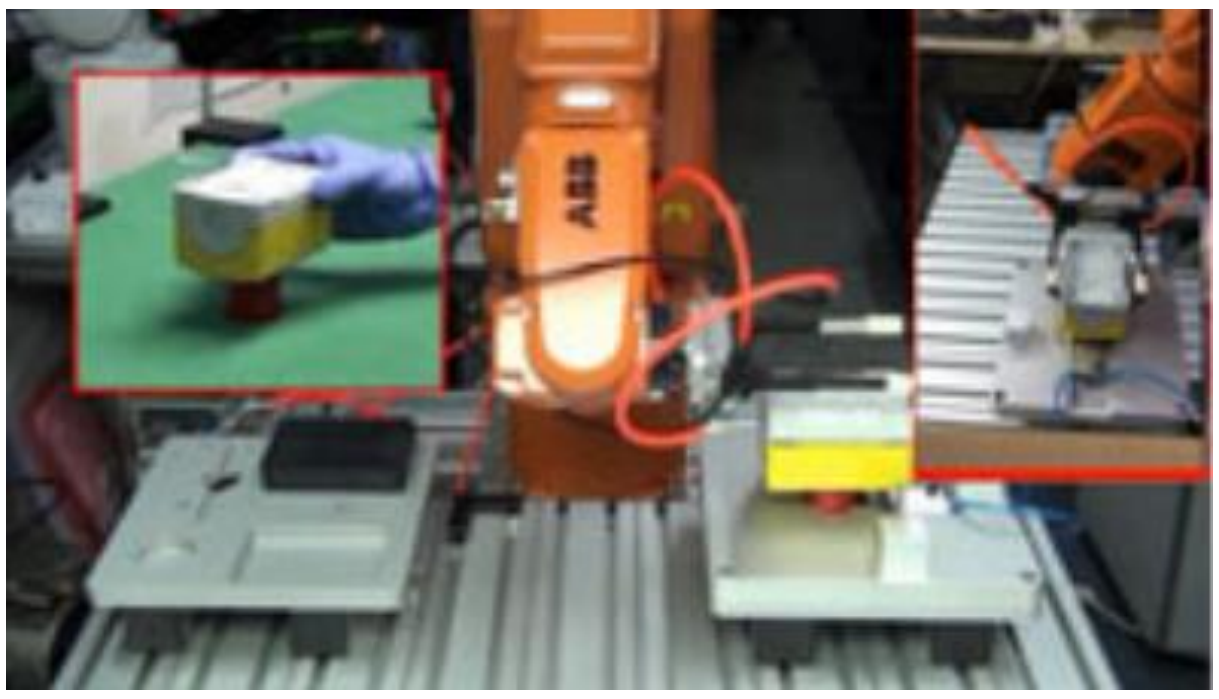

(f) Switch assembly (Yue Wang et al., 2018)

*Figure 3: Practical assembly scenarios*

Related application scenarios do not represent a direct practical application but offer experimental evaluation with objects that are realistically transferable to industrial environments (see Figure 4). A prominent example is the plug-in of standardised connectors, including RJ-45 connectors, USB sticks, power plugs, and HDMI connectors (see Figure 4(a-c)). Specific industries, including the medical, construction and micro-scale assembly sectors, were targeted by (Huang et al., 2022; Aljaz Kramberger et al., 2022; Ma et al., 2022). Experiments demonstrated capabilities for sewing personalised stent grafts whose dimensions were provided by current stent graft manufacturers, constructing timber structures, and performing precision peg insertion tasks (see Figure 4(d-f)). Some researchers chose benchmark models for assembly-associated tasks to evaluate their proposed method (see Figure 4(g,h)). These include the U.S. National Institute of Standards and Technology assembly board #3 (NIST, 2018) and the Cranfield benchmark model (Collins et al., 1985). The latter has been primarily used for peg insertion capabilities (Abu-Dakka et al., 2014, 2015; Savarimuthu et al., 2018). The remaining 45 studies have used industry-unrelated objects. These include arbitrary toy parts or generic machined and 3D-printed components.

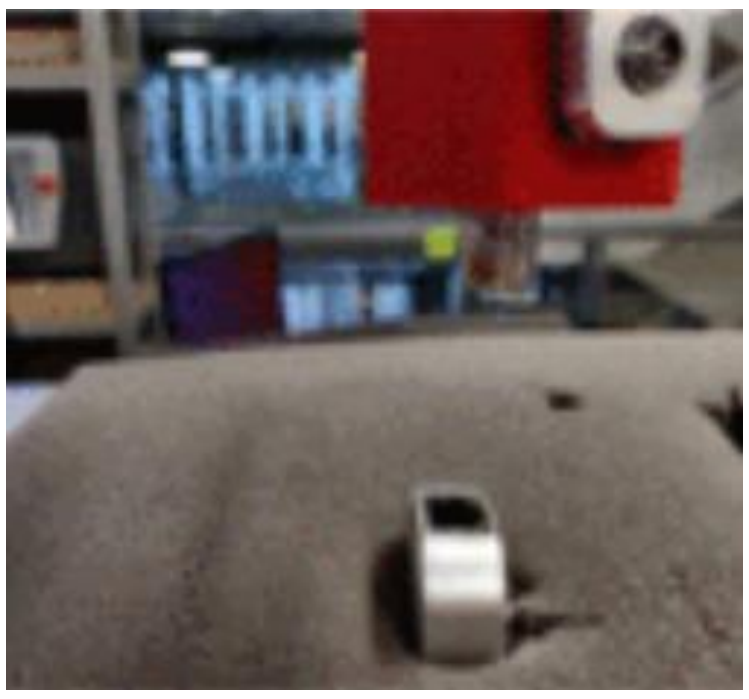

(a) RJ-45 connector (Davchev et al., 2022)

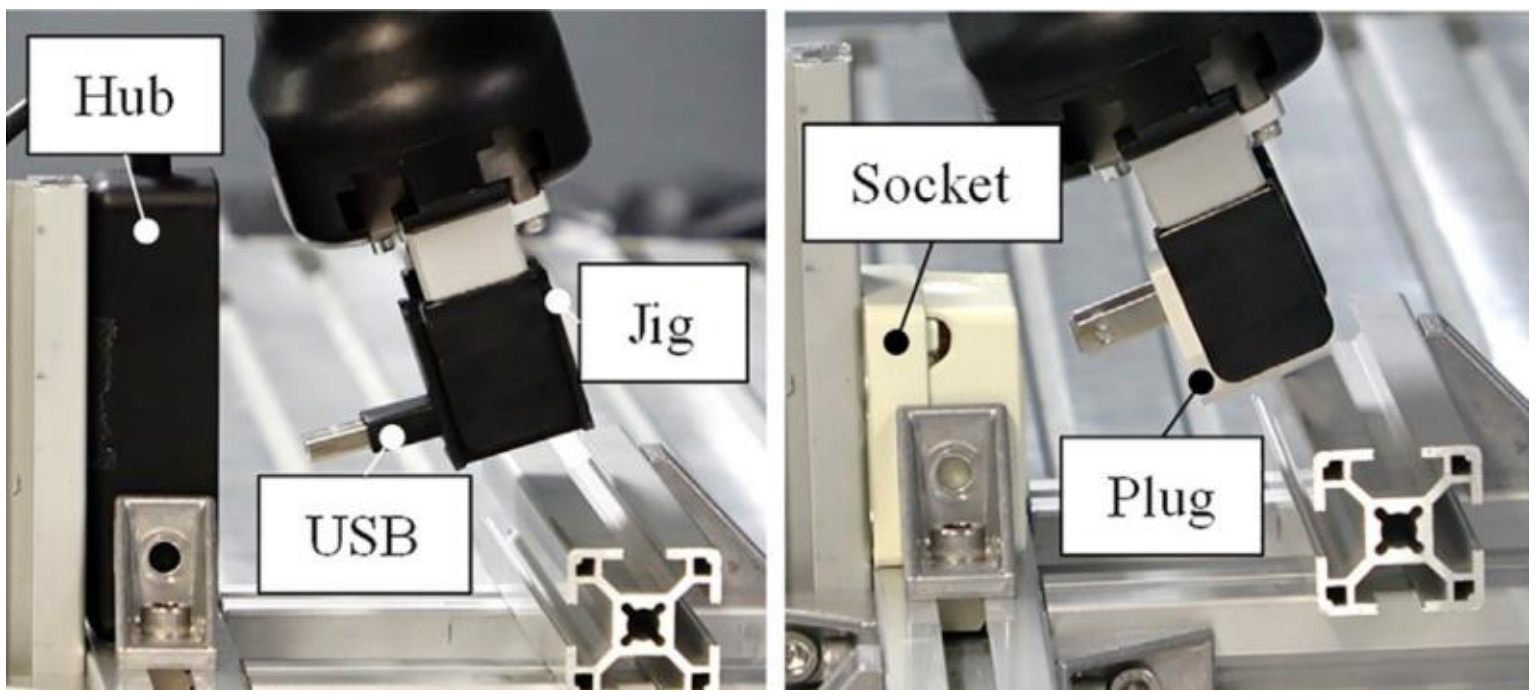

(b) USB stick and power plug (Yan Wang et al., 2022)

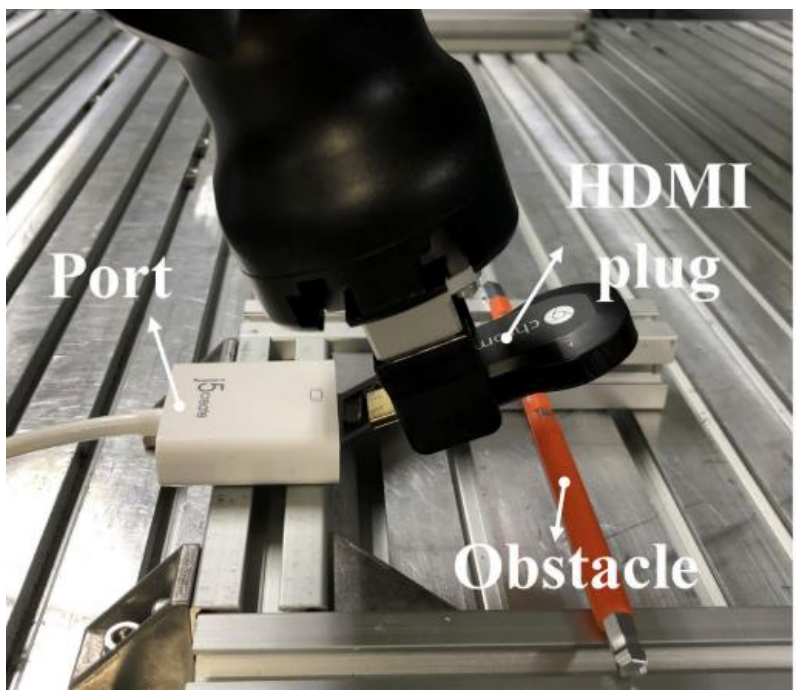

(c) HDMI connector (Yan Wang et al., 2021a, 2021b)

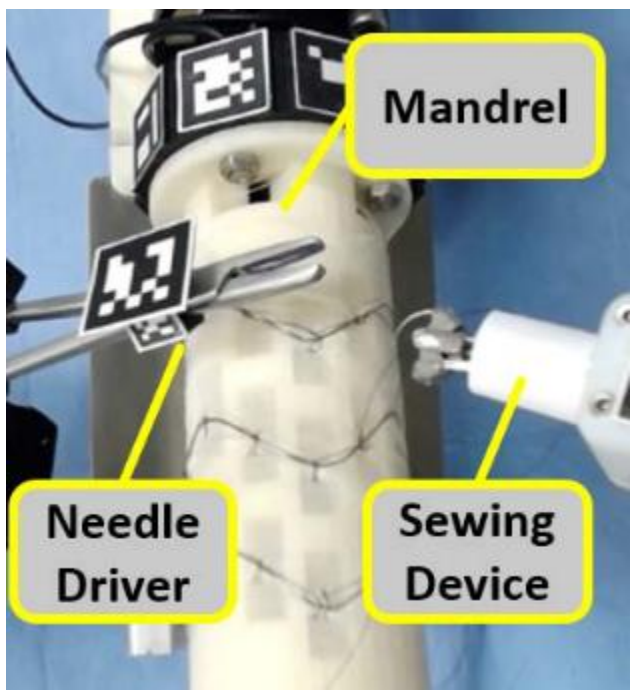

(d) personalised stent grafts (Huang et al., 2022)

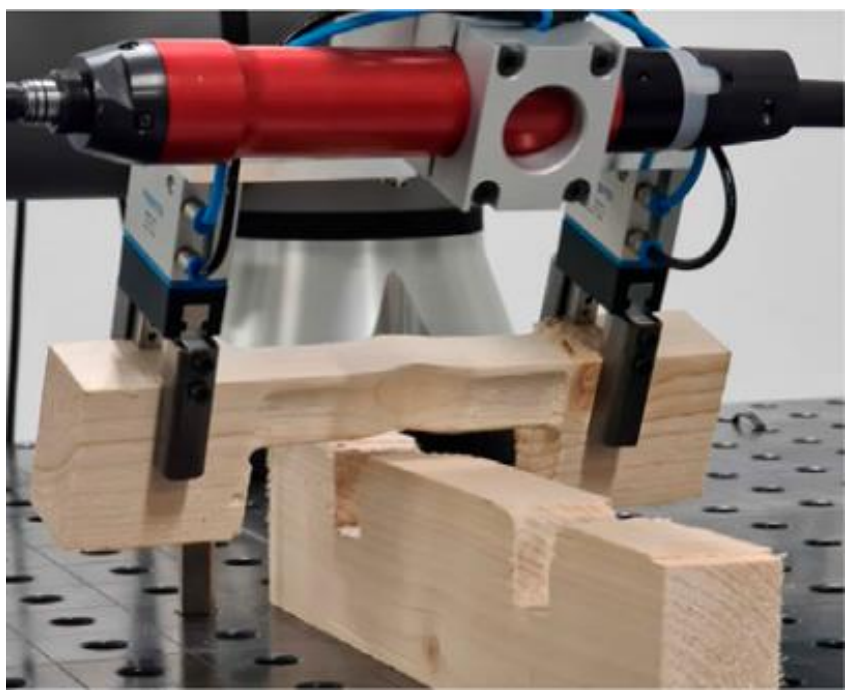

(e) timber structure assembly (Aljaz Kramberger et al., 2022)

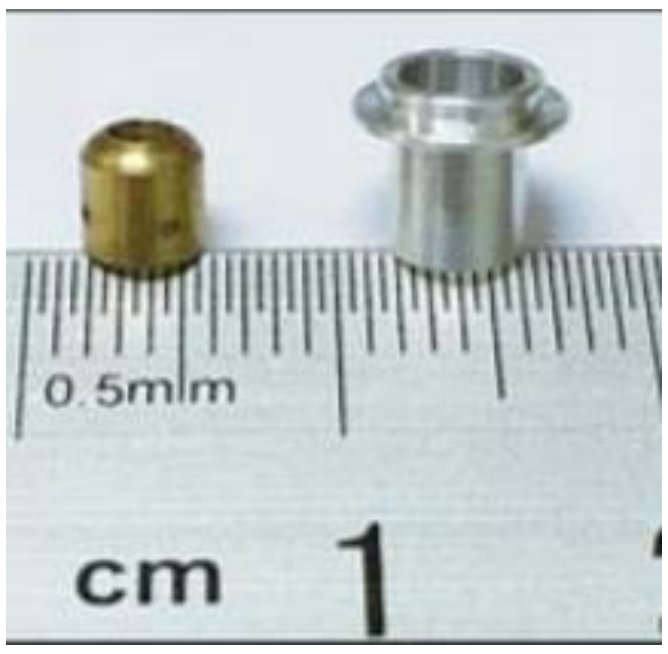

(f) micro-scale peg insertion (Ma et al., 2022)

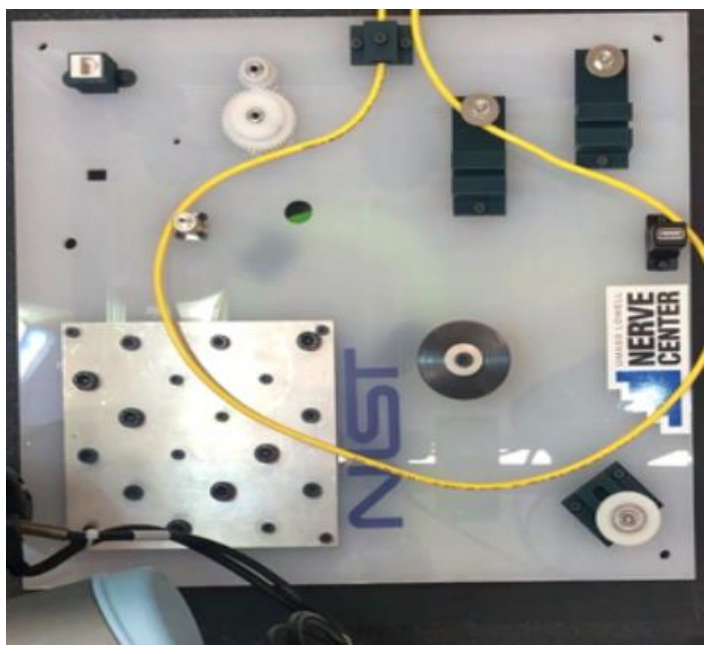

(g) NIST assembly board #3 (Keipour et al., 2022)

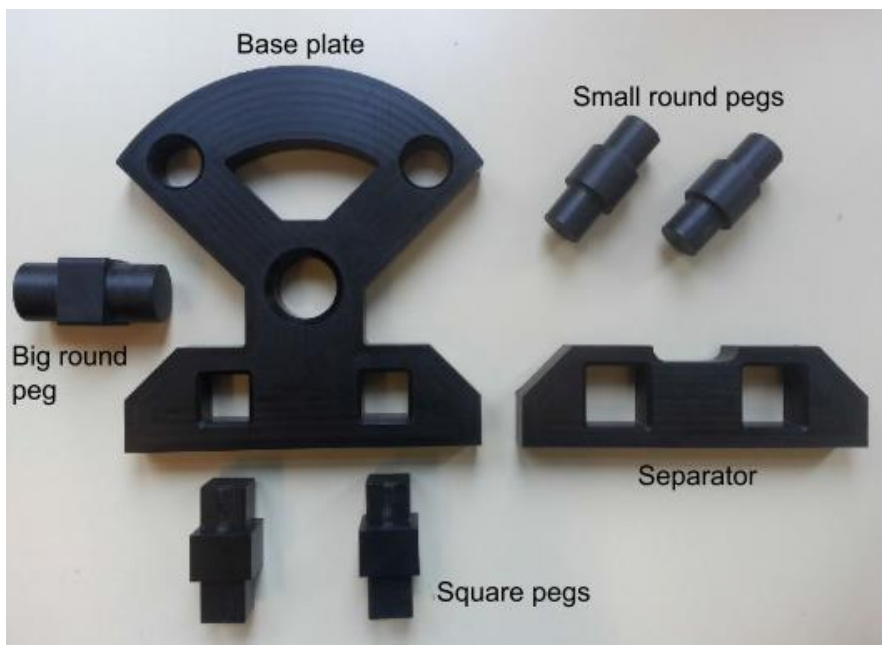

(h) Cranfield benchmark (Abu-Dakka et al., 2015)

*Figure 4: Related assembly scenarios*

Outlining industrial relevance, some efforts are worth mentioning owing to their outstanding experimental setups. As illustrated in Figure 5(a), (Aljaz Kramberger et al., 2022) proposed an LfD* platform in which so-called teaching and execution cells were separated, leading to increased execution space and improved productivity. (Huang et al., 2022) extended the reachability of two surrounding robotic serial arms through an actuated assembly base, allowing to perform sewing motions on all sides of the object (see Figure 5(b)). Challenged by micro-scale assembly, (Ma et al., 2022) emphasised LfD* precision capabilities and built a setup incorporating a three translational degree-of-freedom manipulator achieving a resolution of $1\ \mu m$. Furthermore, the three rotational degree-of-freedom platform was equipped with a force-torque sensor that reaches a force resolution around $1/128\ N$ and two microscopic cameras with zoom lenses were surrounding the workspace to precisely measure the component's poses. A similar system was proposed by (Qin et al., 2019).

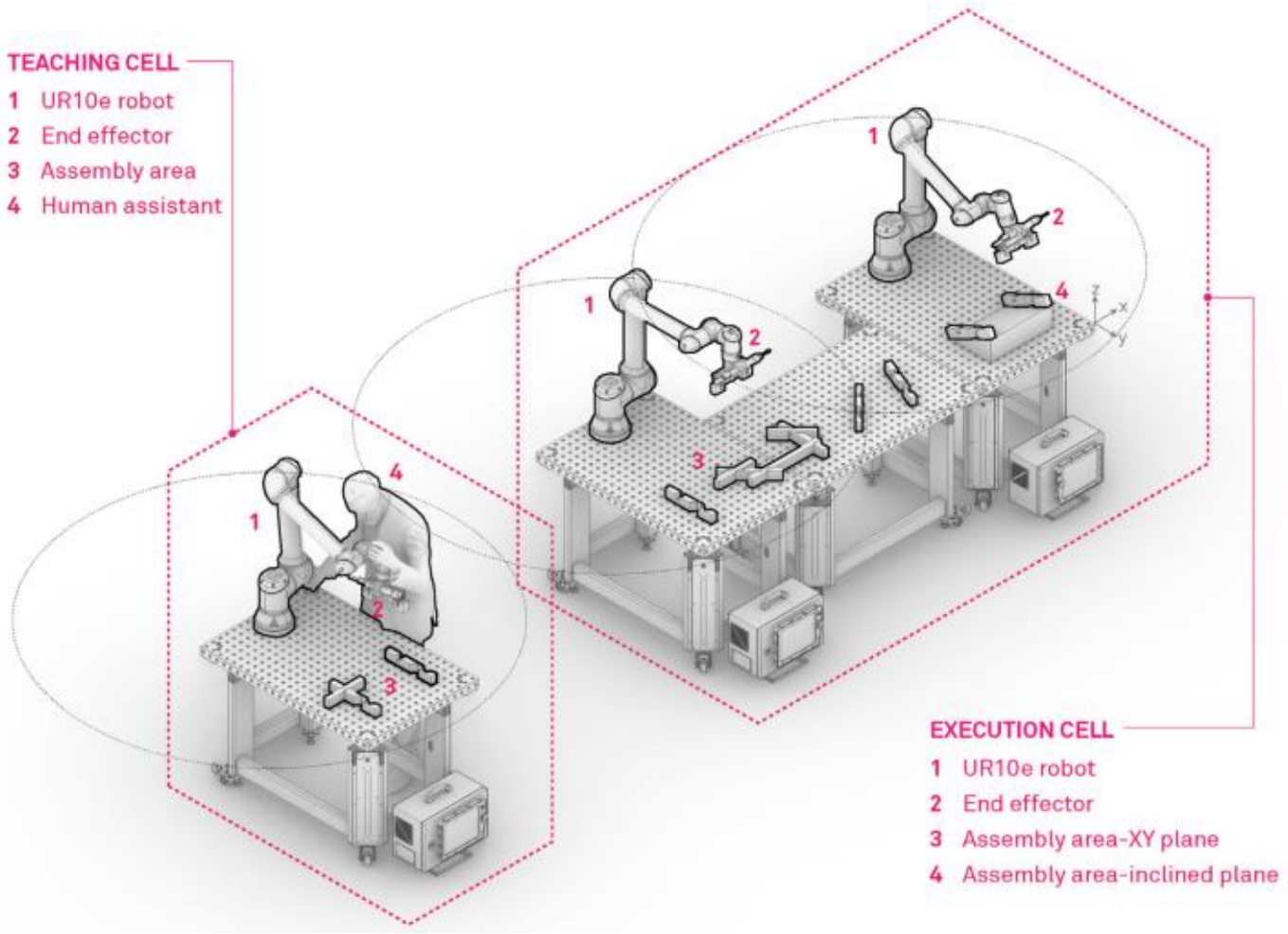

(a) teaching and execution cell separation (Aljaz Kramberger et al., 2022)

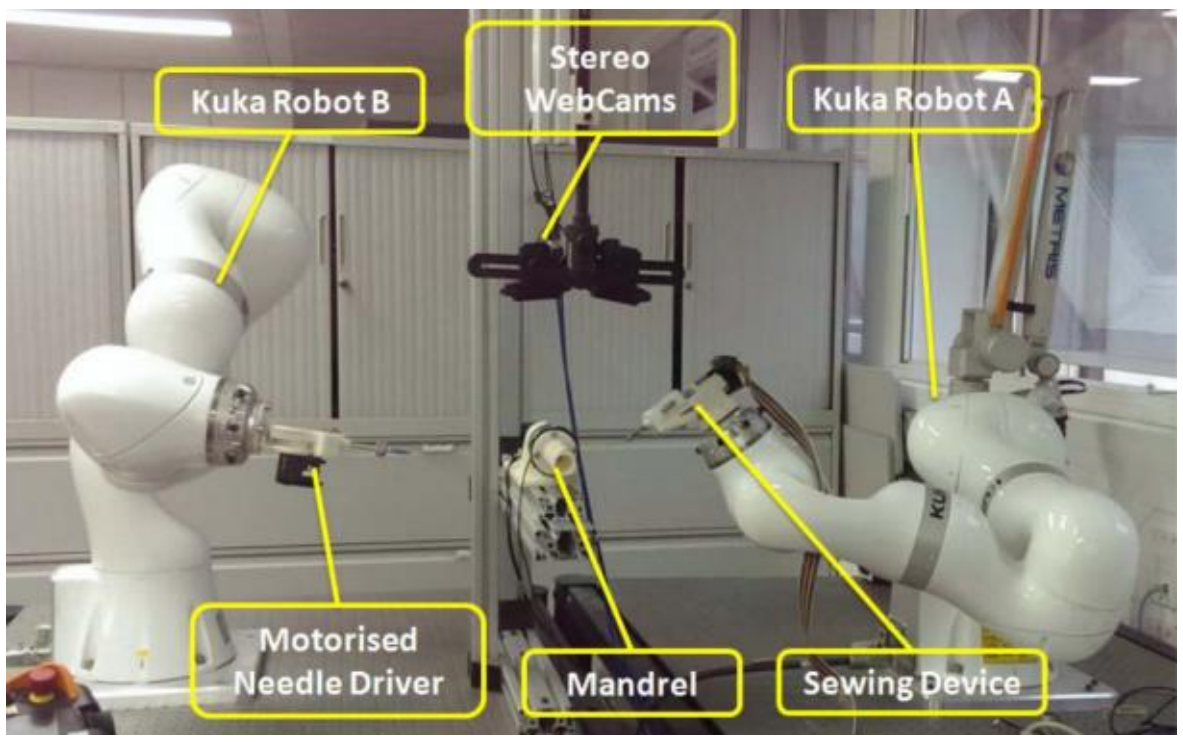

(b) Extended reachability through rotating objects (Huang et al., 2022)

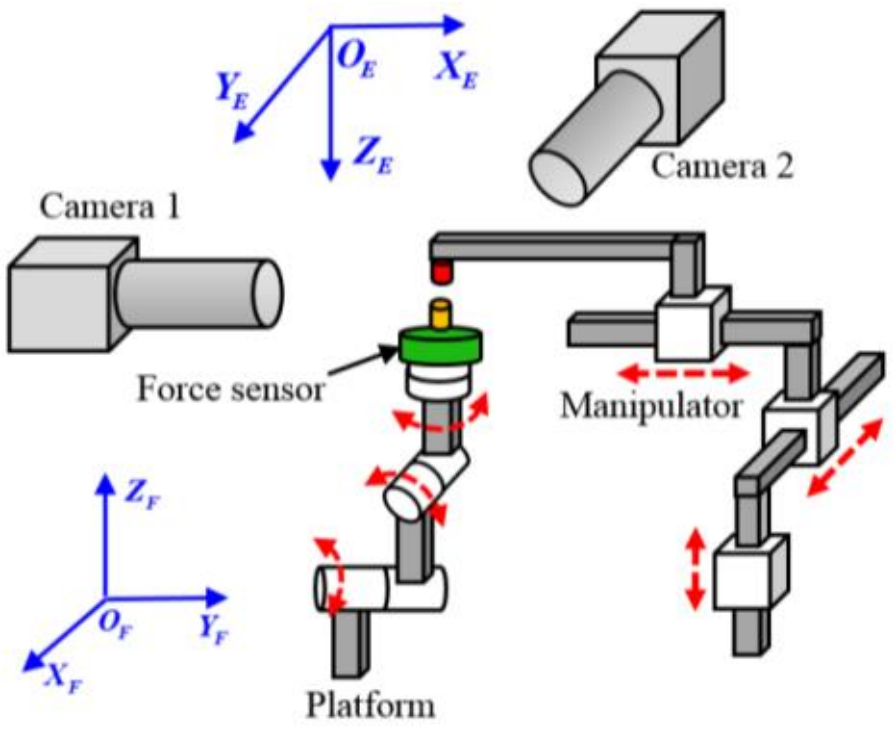

(c) micro-scale precision assembly (Ma et al., 2022)

*Figure 5: Outstanding experimental setups*

### 3.1.3 Individual Assembly Skills

As pursued in the applied search strategy (see Section 2.1), various subskills are considered associated with assembly-related skills. The collection of 61 analysed studies provides experimental evaluations on a total of 77 individual assembly skill references. Figure 6 illustrates the distribution of experimental evidence over potential assembly skills categorised according to mating and joining capabilities as proposed by (Nof et al., 1997). Furthermore, Figure 6 depicts the allocation to the discussed practicability level of Section 3.1.2.

As can be seen in Figure 6, the skill of peg insertion under tight tolerances attains exceptional dominance in academic research. Overall, 46.8% of all experimental scenarios performed this specific mating skill, followed by general pick-and-place skills (loose fitting requirements), covering 23.4% of all reported evaluations. Other mating skills are stacking and bin picking/sorting, which were considered in 7 and 3 use cases, respectively. Joining capabilities are significantly less explored with only a 16.9% share of the 77 discovered assembly skills. LfD* methods were evaluated on screwing, bolting, gluing, wiring, hammering, interlocking, and sewing capabilities. No evidence on the remaining considered joining skills has been reported in the studies, including jamming, riveting, fastening, welding, retaining, press or snap fitting, and crimping.

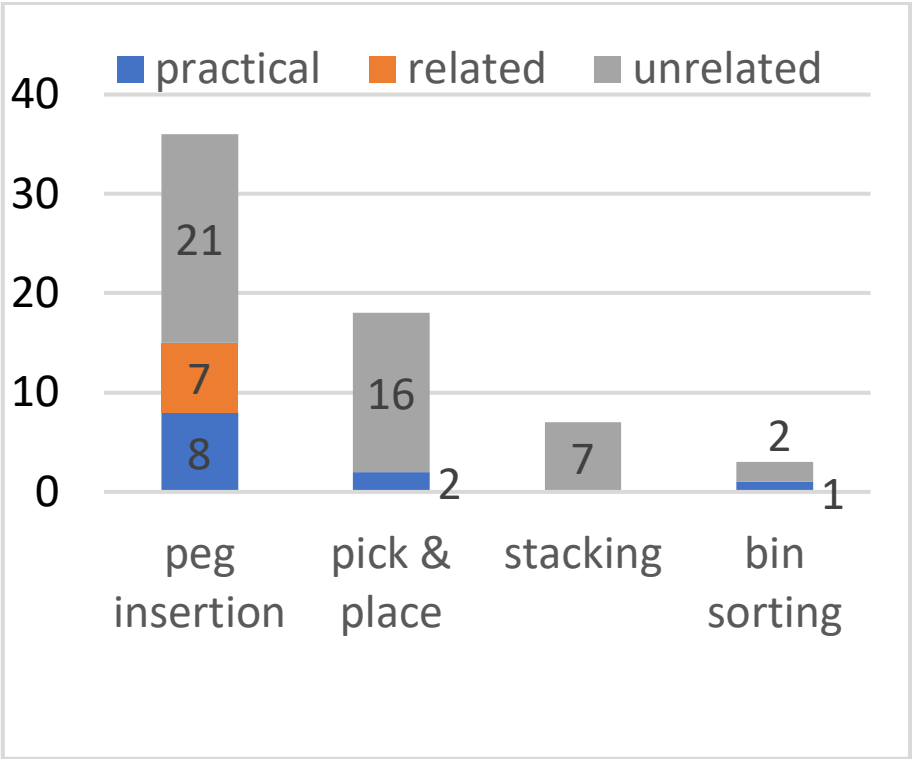

(a) mating skills

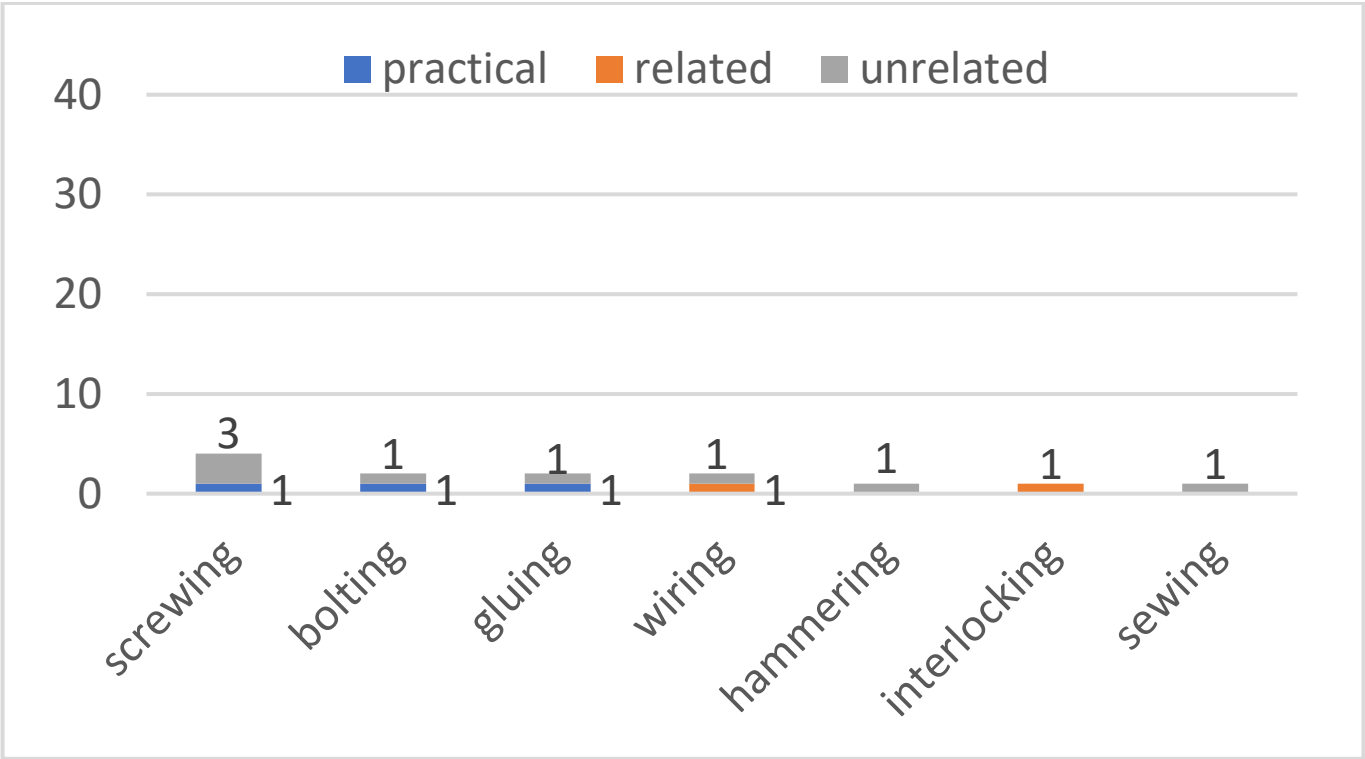

(b) joining skills

*Figure 6: Classification of assembly scenario according to investigated assembly subskills after* (Nof et al., 1997) *and their practicability level*

### 3.1.4 Generalisability

The capability of generalising and expanding to unseen scenarios is one of the key aspects distinguishing LfD* concepts from intuitive programming techniques and promises to enable robotic systems to deal with dynamic environments and product variations in industrial settings. In the context of assembly-related application scenarios, six major generalisation capabilities were identified that have been explored in the reviewed studies. These include the ability to reproduce the task under distinct spatial or temporal requirements (execution scaled to demonstration), cope with task uncertainties, adjust the path or skill sequence, and execute the task with objects similar but not equal to the one used for demonstration (transferability). Note that the following quantification only reflects capabilities that have been evaluated in physical experiments and do not necessarily show all capabilities of the underlying learning method.

With 46 reported evaluations, spatial scaling is the most common generalisation capability explored in LfD* methods, incorporating distinguished start and goal poses for trajectories and object positions. While assuming to know the theoretical location of the objects, the existence of task uncertainties has mainly driven the field of LfD* methods applied to peg insertion tasks. Considering variable grasps of the peg, unprecise locations of the hole, and unmodeled manufacturing defects of the involved components (Shetty et al., 2022) as potential causes of slight deviations to the assumed poses, 18 experimental reports have contributed to counteracting measures. Commonly tackled with Reinforcement Learning techniques, path optimisation was mentioned in eight applications to improve the robots' execution, while graph-based sequence optimisation was only explored twice in experimental reports (Y. Chen et al., 2022; Guo & Burger, 2022). Even though an equally intrinsic feature as spatial scaling of LfD* learning methods, temporal scaling capabilities are often neglected, reaching two mentions in the reviewed studies (Davchev et al., 2022; Gašpar et al., 2018).

Generalising over similar objects has mainly emerged as investigated capability in most recent studies (see Appendix A). Experimental contributions examined similar objects with distinguished properties (Ma et al., 2022; Wan et al., 2017), objects of the same task class but distinguished shapes (Ahn et al., 2023; Cho et al., 2020; Davchev et al., 2022; Yan Wang et al., 2022) or entirely unknown objects (Berscheid et al., 2020; Meszaros et al., 2022). Depending on the learning method, the LfD* approach may not necessarily require any generalisation process (Ahn et al., 2023; Cho et al., 2020), only a few update steps (Davchev et al., 2022) or interactive adaptation (Meszaros et al., 2022). In general, adapting existing knowledge to slightly distinguished situations/objects is promoted with less required effort than demonstrating the task from scratch.

#### 3.1.5 Performance Evaluation

To provide a valid experimental evaluation of a proposed method, different performance metrics were utilised in the reviewed studies. These include, in particular, the reporting of the success of a task or its success rate over several attempts, accuracy analyses, effectiveness compared to competing approaches and achievable efficiency.

While all studies present at least one successful physical execution, a total of 41 studies report achieved success rates over at least three attempts (seven have not specified the number of attempts used to determine the success rate). The reported success rate over the number of attempts is visualised in Figure 7 using the categorisation of mating and joining skills. As can be seen, mating skills tend to be evaluated using more attempts and generally achieve higher success rates.

In the special case of peg insertion, success was often challenged by tight tolerances that are required to overcome. In total, 26 peg insertion skills were assessed with specified tolerances (ten did not provide specifications). In general, tolerances between $0.006\ mm$ (Gubbi et al., 2020) and $6\ mm$ (Jha et al., 2022) with an average of $0.708\ mm$ were considered. Additionally, a few studies chose specifications according to the ISO 286 standard (N. Liu et al., 2020; Tang et al., 2016), inference or clearance fit (Ma et al., 2022; Qin et al., 2019) or a $1\ mm$ chamfer of the hole (Tang et al., 2016). Further accuracy analyses regarding other skill classes reflect the analysis of trajectory deviations from the demonstrated motion (Deng et al., 2022; X. Zhang et al., 2021) as well as the final pose errors in tasks such as pose alignment (Qin et al., 2019; X. Zhang et al., 2021) or placing (Berscheid et al., 2020).

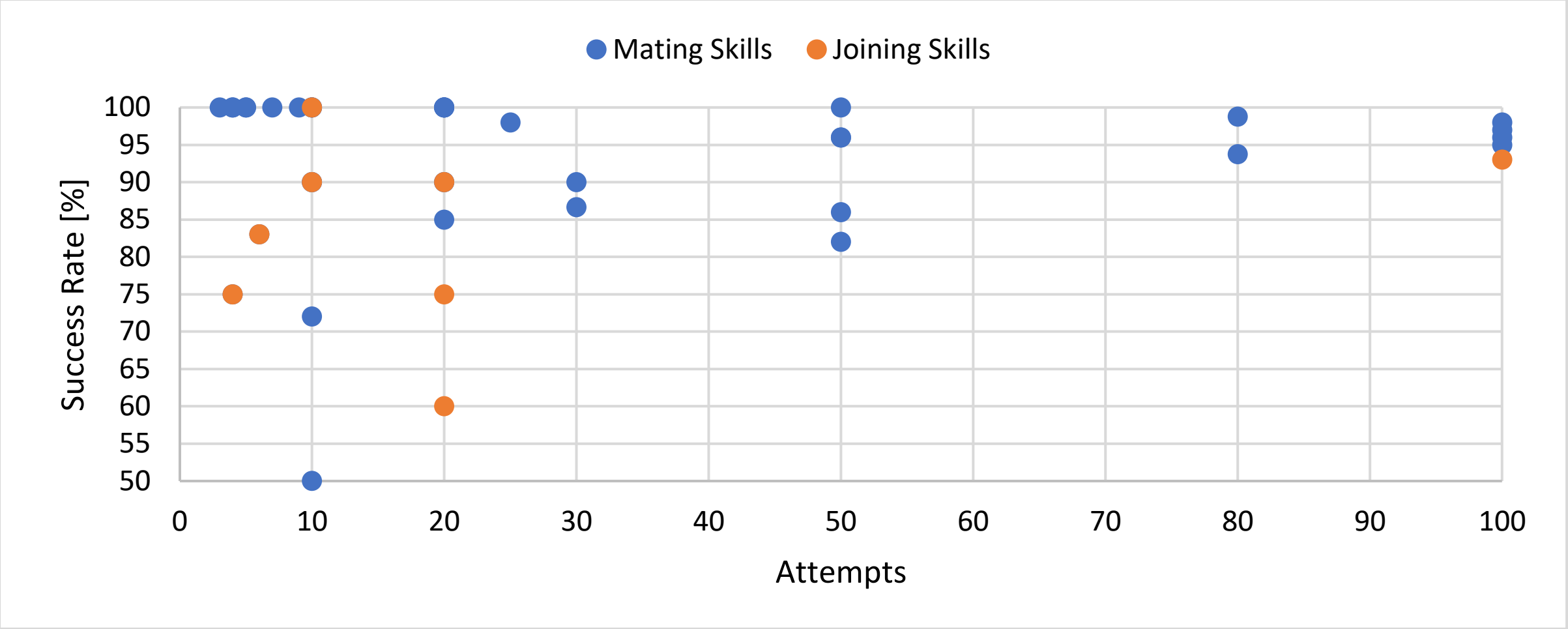

*Figure 7: Reported success rates over attempts distinguished between mating and joining skill categories*

Building on a profound background on LfD* research, a common evaluation practice is given by evidence of the proposed method's effectiveness in comparison to competitive approaches. This can be the ground truth for automated recognition (Eiband et al., 2023; Gu et al., 2018), comparable techniques (Aljaž Kramberger et al., 2017; D. Liu et al., 2022; Hongmin Wu et al., 2022), baseline method when extended (Davchev et al., 2022; Yan et al., 2023), or the comparison to other representative LfD* approaches (Ahn et al., 2023; Jha et al., 2022).

In addition, some studies reported performance assessments towards procedure efficiency. Besides reports on a reduced number of required execution steps (Yan Wang et al., 2022) or sequenced actions (Y. Chen et al., 2022), evidence of improved execution speed is provided by several researchers (Abu-Dakka et al., 2015; Cho et al., 2020; Davchev et al., 2022; Aljaž Kramberger et al., 2017; Meszaros et al., 2022). Through interactive correction techniques, (Meszaros et al., 2022) extends the efficiency assessment to the training time, reporting on a four times faster approach compared to the initial demonstration method. Similar demonstration time reduction efforts were promoted by (D. Liu et al., 2022; Hongmin Wu et al., 2022). (Yan Wang et al., 2021a) used pretraining in simulation to reduce the time required on the real robot. As the reduction of computing time is equally important for improved efficiency, (Y. Q. Wang et al., 2021) optimised the underlying learning technique, and (Keipour et al., 2022) developed a routing method efficient enough to run online. On the other hand, (Xu et al., 2022) raised performance concerns regarding the proposed RL-centred method when confronted with more complicated tasks. Targeting 3C product assembly operations with optical motion capture, (Hu et al., 2021) proposed a performance evaluation protocol including performance indicators for static and trajectory evaluations.

## 3.2 Qualitative Results

The qualitative analysis of the identified literature provides insights into the current state of research on assembly-related LfD* methods with an emphasis on targeted challenges, methodological approaches, and achievements. The analysis is structured according to individual assembly skills (see Figure 6), grouped into peg insertion, pick-and-place-related and joining skills. Due to their similarities, we consider reaching, grasping, stacking and bin sorting/picking as special cases of pick-and-place-related skills.

### 3.2.1 Peg Insertion Skills

The ability to insert an object, in this context colloquially known as a "peg", into a dedicated hole is a prevalent and essential task within the manufacturing industry (Cho et al., 2020). As typical assembly operations include many peg-in-hole tasks (Arguz et al., 2022), it is considered one of the standard operations required to be solved for automated assembly (Abu-Dakka et al., 2014). Although seemingly simple in nature, the challenge lies particularly in the high precision requirements often demanded due to tight clearances between the objects or occurring task uncertainties (see Section 3.1.5). The research interest in this particular skill is reflected in the dominant quantity of 36 out of 77 reported experimental evaluations.

Considering task uncertainties, several studies investigated the use of compliant controllers, including variations of admittance (Abu-Dakka et al., 2014; Qin et al., 2019; Savarimuthu et al., 2018; Stepputtis et al., 2022) and impedance (Abu-Dakka et al., 2015; Gubbi et al., 2020; Aljaž Kramberger et al., 2017; N. Liu et al., 2020; Su et al., 2022) control methods, that allow an adjustment of the robot motion based on measured contact forces. While some contributions learned from a single demonstration, e.g. (Abu-Dakka et al., 2014), trajectory-based methods show a decrease in success rate when the spatial parameters of the execution deviate significantly from the demonstration poses (Arguz et al., 2022). Therefore, others proposed the consideration of multiple demonstrations to create generalised (Aljaž Kramberger et al., 2017) or prescribed (Su et al., 2022) force profiles covering a broader spectrum of potential conditions. Furthermore, (Yan Wang et al., 2021b, 2021a) developed a Reinforcement Learning-based compliant control policy in combination with nominal motion trajectory through a hierarchical imitation learning framework, while (Gubbi et al., 2020) investigated the applicability of the recently developed generative adversarial imitation learning approach on high-precision peg insertion. The idea of compliant controllers was furthermore evaluated on robot systems with reduced capabilities, including industrial robots with non-backdrivable mechanisms and strict tolerance requirements (Tang et al., 2016), position-controlled industrial robots (Jha et al., 2022), or even without force sensor using ergodic exploration (Shetty et al., 2022). Beyond the dominant case of single peg insertion, (Ma et al., 2022; Stepputtis et al., 2022; Yan et al., 2023) achieved multi-peg-in-hole assembly operations by reacting on force measurements.

While emphasising the specific assembly skill of peg insertion, some studies investigated strategic approaches to fulfil the task successfully. (Wan et al., 2017) suggested the calculation of an optimal path by reducing the demonstrated motion mathematically to the axis alignment. (Ti et al., 2022) defined an intermediate three-point contact state with an assembly-angle representation for round peg insertion tasks. An exception strategy was proposed by (Abu-Dakka et al., 2014) for the systematic search of the hole by initially starting with deterministic translational variation followed by a stochastic search with random increments. In case of misalignment or external perturbation, (Stepputtis et al., 2022) combined a phase estimator with an admittance controller to enable the robot to correct or even reverse the task progress. Considering a differentiation of errors parallel and perpendicular to the assembly surface, (Ahn et al., 2023) developed two separate trajectory generators to respond to alignment and insertion processes.

Motivated by the idea of minimal demonstration input, (Berscheid et al., 2020) proposed a method that learns from goal state images allowing them to succeed in generic pick-and-place as well as peg insertion tasks with $1\,mm$ tolerance. Evaluated comprehensively on practical assembly tasks incorporating peg insertion, (bin-) picking, and placing skills, (Ji et al., 2021) used similarly passive observation to reduce the human effort for automating robotic assembly. Based on the extracted assembly sequence, state transitions, grasping modes and involved objects, the proposed framework automatically generates a robot assembly script considering the different embodiments by utilising pre-trained robot skills, self-exploration, self-reproduction, and self-improvement capabilities. Such self-driven learning after the initial human demonstration is common practice for LfD* methods based on Reinforcement Learning (RL) techniques. While defining distinctive primitive skill libraries for the hole search and peg insertion, (Cho et al., 2020) used RL to optimise the generated motion based on either previously experienced or newly defined skill instances. (Davchev et al., 2022) applied model-free RL to learn a residual correction policy. The RL-based controller by (Yan Wang et al., 2022) is capable of learning the control policies of a specific class of complex contact-rich insertion tasks based on the trajectory profile of a single instance that enables the generalisation to similar objects.

Considering the cost and burden on the human operator of each demonstration, (Ma et al., 2022) artificially augmented the number of demonstrations reducing the required number to one-third for the consecutive RL-based self-learning assembly phase. While studies emphasising the above capabilities assume the existence of perfect demonstrations, (Pervez et al., 2017) explored the situation in which the operator may fail to provide multiple complete demonstrations. The developed stochastic model allows the execution of the given task based on one full demonstration and multiple incomplete/inconsistent attempts.

In light of extended assembly capabilities, such as general pick-and-place motions, a few studies outline methods beyond the contact-rich peg insertion task. The distinction motivated the definition of distinguished image-featured guided and force-constrained motions (Qin et al., 2019) or approaching and assembling motions (Hu et al., 2022). The latter was created based on the idea that within the industrial field, the generalisation ability to environmental constraints is of more importance during the approaching phase, while object constraints matter during the assembly phase. (Duque et al., 2019) investigated the applicability of Petri nets to automatically generate work plans according to available objects within the workspace and generalise to new scenarios. Finally, (Savarimuthu et al., 2018) created a sophisticated three-level architecture that extends the adaptation of sensorimotor skills (Abu-Dakka et al., 2015) with key-frame-based semantic and pre- and post-conditioned planning levels. Additional techniques were incorporated for self-learning and human interaction for efficient decision-making.

### 3.2.2 Pick-and-place-related Skills

The skill of picking up an object from one location and placing it somewhere else is arguably the most widely used skill in robotics within general manipulation tasks. Considered a generic skill, it is essential for the success of assembly-related tasks as these usually consist of multiple sequentially connected subtasks to handle multiple objects. Pick-and-place-related skills, including reaching, grasping, stacking and bin sorting/picking, differ from specific peg insertion in two essential aspects. It generally requires less strict tolerances to be maintained, and the aimed outcome can often be achieved through distinctive action sequences (path alteration, object handling sequence, etc.), contrary to the deterministic success characteristic of peg insertion. Identified as an essential component for performing assembly-related tasks, 28 out of 64 reported mating skill evaluations emphasised on pick-and-place-related characteristics (see Section 3.1.3).

In line with the above-discussed approaches to extending peg insertion capabilities (see Section 3.2.1), a common approach to tackling motions beyond contact-rich skills is the development of appropriate subskills or subgoals that are separately learned and joined afterwards to achieve the expected assembly outcome. (Sefidgar et al., 2017) developed a tangible programming technique with predefined objects for indicating subskills that translate to a sequence of robot functions, including instances of *pick-up-from-top*, *pick-up-from-side*, *place-at* and *drop*. In contrast to this deterministic approach, (Yue Wang et al., 2018) proposed an automatic programming method for robotic assembly that estimates the present assembly skill and involved parts from a recorded video segment. This framework distinguished between the predefined skills of *placing*, *screwing*, *taking*, *pushing*, and *labelling*. In the context of stacking capabilities, (Kang & Oh, 2022) defined *reaching*, *picking*, *carrying*, and *placing* as base skills and argued that skills connecting those are challenging to obtain via demonstrations due to their arbitrary. Hence, the proposed base skills were acquired using expert demonstrations, while bridge skills were trained through Reinforcement Learning. A similar idea was proposed by (Pinosky et al., 2022), where actions were artificially synthesised when the policy was uncertain, i.e. regions where the expert demonstration was lacking. A more abstract approach was pursued by (D. Liu et al., 2022), who modelled manipulation tasks as a series of *what-where-how* elements, reducing the attention to the selected object and action for improved adaptability.

An alternative stream of research efforts investigated the automatic extraction of the required action sequence based on determined keyframes (Haage et al., 2017; Perez-D'Arpino & Shah, 2017), goal images (Berscheid et al., 2020; Hongtao Wu et al., 2022), positions of interests (Pellois & Brüls, 2022), and key hand points (Deng et al., 2022). In order to capture the complexity and possible transitions performing multi-step assembly tasks, (Y. Chen et al., 2022) developed a universal functional, object-oriented network that optimised the assembly sequence from multiple demonstrations. Similarly, (Guo & Burger, 2022) proposed a framework evaluated on an inspection and bin-sorting task that established coordination schemes to select the correct sequence of skill primitives, ensuring an appropriate grasping orientation. (Hongtao Wu et al., 2022) achieved zero-shot generalisation to unseen tasks through a novel method of rearrangement from image data, while (Eiband et al., 2023) specialised in an automated segmentation method of trajectory data into *logical* and *classified* skills. Those were implemented using symbolic pre/post-conditional recognition and data-driven sliding windows, respectively.

In addition to the above fundamental concepts for generic pick-and-place-related skills, several studies target specific capabilities that promise valuable contributions to the robustness of LfD* approaches in practical environments. Of particular interest is the secured performance in dynamic situations. (Ghalamzan E. & Ragaglia, 2018) emphasised the dynamic work environment in which the robot was capable of avoiding collision with moving obstacles. Motivated by reducing the execution time, (Meszaros et al., 2022) investigated an interactive correction method to iteratively speed up the non-zero-velocity picking skills of objects. (Y. Q. Wang et al., 2021) optimised an LfD* technique to eliminate errors from human demonstrations by smoothening reproduced motions in appropriate segments. To improve the robustness of the robot's performance, (Iovino et al., 2022) introduced additional verbal interaction to clarify potential disambiguation in the scene, e.g. when identical objects are present in the workspace, and (Hongmin Wu et al., 2022) developed a method that enabled the robot to quantify its learning progress and guide the user to efficient demonstrations. Incorporating external forces, (Y. Zhang et al., 2022) proposed a method that reduced the impact of external disturbances, which was demonstrated by the example of task completion despite physical interaction with the robot arm. Alternatively, (Ugur & Girgin, 2020) suggested the use of external forces in uncertain situations to manually adjust the ongoing movement by physical interaction with the robot.

### 3.2.3 Joining Skills

In a general sense, the assembly process refers to the superimposed steps of mating and subsequent fastening of components. Joining skills are usually an inevitable necessity in order to connect assembly components robustly and ensure the designated function of the final product robustly or permanently. Nevertheless, as discussed in Section 3.1.3, most studies focused on the implementation of targeted mating skills in the LfD* context, resulting in only 13 out of 77 reported experimental investigations towards joining skills.

As indicated in Figure 6, *screwing* was the most investigated joining skill in the reviewed studies with four experimental evaluations. (Ji et al., 2021) and (Yue Wang et al., 2018) developed comprehensive assembly systems, both utilising passive observation with consecutive assembly skill estimation, part recognition and robot embodiment strategies, capable of performing distinctive assembly skills, including *bolting* and *screwing*, respectively. The experimental evaluations provide limited insights into their performance, apart from mentioning failed attempts (Yue Wang et al., 2018) and identified issues attributed to insertion tolerances (Ji et al., 2021). Evaluated on toy components, (Gu et al., 2018) assessed an assembly sequence including *bolting*, *hammering* and *screwing* skills through passive observation of human performances. To reach the *screwing* assembly state, eight turns were required. The repetitive turning characteristic was tackled by two identical markers on the screw that enabled 180-degree rotations to be performed in either configuration. In this setup, *screwing* was identified as the most challenging skill due to the complex motion in addition to small-sized screw slots, while *hammering* was considered the simplest task due to the simplicity of motion and loose hitting accuracy. Emphasising unstructured demonstrations, (Niekum et al., 2015) proposed a method for automatic skill segmentation and interactive human intervention. Evaluated on the assembly of table legs with protruding screws, the system performed the *screwing* task after interactive correction using recovery behaviour for difficult grasping angles or distant leg locations. Based on a skill library including *screwing* (clockwise + anti-clockwise) and *stacking*, (Yu & Chang, 2022) proposed an RL-centred method that maps new scenarios to a sequence of a few library instances. The evaluation was conducted on a simplified task design with loose clearances and no contact with the environment, limiting the practicability assessment of the method for realistic *screwing* tasks.

Motivated by precision assembly challenges, (Qin et al., 2019) demonstrated *gluing* capabilities on a sleeve-cylinder assembly task after insertion using a predefined force-constrained motion action class. As described in Section 3.2.2, (Eiband et al., 2023) focused on the automatic segmentation of the demonstration data where the *gluing* motion was abstracted to a sliding skill with slight pressure against the surface. The experimental *gluing* motion was performed under loose spatial or task-related requirements.

In the case of *wiring*, it requires the handling and routing of deformable wires. (Keipour et al., 2022) developed a spatial representation graph that enables the rerouting of wires, considered a pick-and-place task, towards a goal configuration. Emphasising the enhancement of kinesthetic teaching through online impedance shaping, (Meattini et al., 2022) evaluated its method on a *wiring* task consisting of the pick-up of a cable's extremity and insertion it into a connector. The execution behaviour was furthermore altered at run-time through physical interaction to perform a rerouting of the cable.

While *sewing* represents a highly repetitive task and is considered the most challenging skill in the manufacturing of personalised stent grafts, (Huang et al., 2022) developed a robotic system capable of extending a single demonstrated stitch cycle to the whole stitching task through design specifications, rigid transformations and an actuated mandrel. (Aljaz Kramberger et al., 2022) investigated a robotic *interlocking* method that enables timber-timber joinery without the need for additional steel fasteners through optimal structural truss design and particular rotational insertion strategy.

# 4. Discussion

Building on the results of the systematic literature review providing comprehensive insights into the state-of-the-art of LfD* efforts in assembly-related applications, this Section seeks to answer the research questions RQ3+4. In this context, we believe that mimicking human operators' capabilities regarding learning and executing new tasks provides the highest probability of acceptance for LfD* solutions in the traditional assembly industry. Subsection 4.1 assesses the reported LfD* findings regarding their transferability to industrial practice by comparing them to a well-established instruction method for human operators. Subsection 4.2 outlines identified obstacles in current LfD* research that dictate the discrepancy between research achievements and industry requirements. Finally, Subsection 4.3 discusses the limitations of the conducted systematic literature review process.

## 4.1 Comparison to Industrial Practices

Industrial assembly builds on well-established and historically preserved methodologies for instructing human operators that have proven great effectiveness. In traditional manual industrial assembly, instruction refers to the systematic learning of knowledge and skills to perform a task that operators are expected to carry out in a production environment but have not previously fully known or mastered (Maier et al., 2020). Instruction concepts are a measure of operator qualification and human resource development, and the theoretical concepts are correspondingly diverse. A primary distinction can be made between on-the-job, along-the-job, near-the-job, and off-the-job training (Berthel & Becker, 2022). According to (Schelten, 2005), instruction procedures can be further differentiated depending on the learning domain (sensorimotor, sensorimotor and cognitive, cognitive) as well as the degree of involvement (instructor emphasised, instructor and learner involved, learner emphasised).

As outlined in Section 3, research efforts towards LfD* solutions for assembly-related applications emphasise on-the-job instructions for sensorimotor capabilities. This promotes the comparison of state-of-the-art LfD* solutions with the so-called four-step method, an industrially relevant and well-established method. It is based on the principles of the "training-within-industry" approach (TWI), which was developed in the USA during the Second World War (Jung, 2016) and has been considered in Germany since the 1950s as the "REFA-Vierstufenmethode" for work instruction (Becker et al., 1993; Verband für Arbeitsstudien und Betriebsorganisation, 1991). It focuses on manual, relatively short-cycled, and simplistically structured tasks that are to be performed repetitively according to a standardised sequence, thus enabling a certain degree of transferability to automated practices (Verband für Arbeitsstudien und Betriebsorganisation, 1991). The subsequent discussion highlights identified synergies and discrepancies between the four-step method and LfD* practices (Becker et al., 1993; Schelten, 2005).

**Preparation –** In the first phase of the four-step method, the learning location is prepared for the instruction and a suitable learning objective is defined based on the learner's existing knowledge. Attention must be given to the interpersonal factors that enable the instructor to arouse the learner's interest and motivation.

The reviewed LfD* approaches show, in principle, transferable competencies in dealing with workplace requirements, including varying setups and dynamic environments. The preparation of the learning location is therefore considered comparable. However, effective interaction with the robotic system relies on the instructor's active engagement with the programming method, understanding the system's capabilities, limitations, and expected performance. This shifts the necessary recognition of interpersonal factors to the instructor's interest and motivation, which must be aroused by an adequate design of the robotic system. The emerging interest in advanced generalisability, i.e. the transfer of knowledge and skills to similar objects and situations, is an essential prerequisite for imitating humans' prior knowledge repertoire.

**Demonstration and Explanation –** This step requires a high level of performance from the instructor and involves the physical demonstration of the task. The instructor performs the task while clarifying the reasoning and important details verbally. These help the learner to sharpen the understanding of the assembly task and to achieve high quality during self-performance.

The active demonstration of the task corresponds to the first phase of the common LfD* procedure. The applied demonstration method determines the required involvement level of the robotic system. Using passive observation techniques provides a comparable method to conventional human instruction with no involvement of the physical robot. However, the most common approach of kinesthetic teaching demands the presence of the robot and the instructor to physically move the robot. This highlights a great distinction to conventional instruction techniques and may be perceived as less practical. Furthermore, the four-step method envisages simultaneous physical demonstration and verbal explanation, while most efforts in research rely on a single predefined channel limiting the instructor's communication abilities. An additional distinction emerges from the ability to comprehensively understand the task, as most LfD* approaches target the acquisition of sensorimotor skills. Few investigations provide ideas on communicating the robot's understanding to the instructor, either immediately or subsequently, increasing the instructor's confidence regarding the perceived demonstration.

**Execution and Explanation –** After passively observing the instructor's demonstration of the task, this step involves mimicking the task by the learner and justifying the key aspects in the learner's own words. The instructor observes and intervenes when necessary. An important aspect is to ensure that the learning objectives are met and understood.

Incorporating the transition of the robotic system from a passive observer to an active executor corresponds to the last phase of the common LfD* procedure. Synergies are identified, in particular, regarding the performance of the task on the basis of the robot's own understanding and skills. Several studies have also presented ways of communication between the robot and the instructing human operator during execution to reason the actions performed, communicate the learning progress or visualise the task understanding. This often involves the creation of collaborative situations with intentional interaction that allows the human operator to intervene and improve the understanding of the task until the learning objective is met.

**Practice –** The last step of the four-step method has the purpose of providing the opportunity for the learner to practice the new skill and apply the acquired knowledge. This step is important for solidifying learning, building confidence and becoming more efficient. The goal is to achieve a defined performance under the premise of error-free assembly.

The learning behaviour of humans can generally be illustrated by learning curves (Maier et al., 2020; Ullrich, 1995). These show the learning progress, i.e. the time required per execution, as a function of the number of repetitions. During the process of understanding the required contexts and actions, the curve declines sharply at first. The learning effect then decreases steadily with an increasing number of repetitions until a routine of working is finally developed, and the assembly process is increasingly internalised. The description of the functional relationship of the learning curve is based on the task to be performed as well as a variety of factors that depend on the learner (Jeske, 2013). With regard to the expected efficiency of an experienced operator in performing industrial assembly tasks, the method of methods-time measurement (Maynard et al., 1948) is considered a well-established industrial practice.

In comparison to the required practice in conventional settings, most LfD* approaches provide the ability to generally start at a higher performance level from the beginning. However, due to the strongly narrow cognitive abilities and hardware limitations, most of the proposed LfD* solutions have rather flattened learning curves. The ability to improve their own performance in a self-controlled manner can be traced in machine learning-based LfD* models. Reinforcement Learning, which has been applied for this purpose in various studies, can be named a key enabler. A prediction of the expected efficiency has not been discussed in the reviewed literature.

## 4.2 Current Obstacles in LfD* Research

While several synergies with a well-established instruction technique have been identified that promote reasonable potential, some fundamental aspects are lagging in creating a smooth integration of LfD* solutions into industrial practices. Based on the findings above, the following provides an educated summary of identified obstacles in LfD* research to drive its progression towards meaningful deployment. These include aspects regarding practicability, task complexity and diversity, generalisability, performance evaluation and integration concepts.

**Practicability –** The practical evaluation of promising solutions that have emerged from research marks a pivotal moment in gauging their potential interest for industrial sectors. Demonstrating significant advantageous features, along with robustness to withstand industrial conditions, instils a willingness to invest in the proposed technologies.

The academic advancements in the field of LfD* over the course of the last decades exhibit a robust level of maturity within the research landscape. State-of-the-art studies successfully demonstrate the robustness and features of popular techniques and increasingly expand their interest in niche challenges. However, the analysis of practicability reveals that only very few studies have presented the application of proposed LfD* methods in practical or related task designs resulting in a severe limitation on evaluations using mainly unrelated tasks and objects. Moreover, the identified practical and related application scenarios can be characterised as tasks which are predisposed for automation. Exploring the potential expansion of a robot's functionalities into tasks that currently fall outside the scope of automation, particularly those involving assembly situations primarily carried out by human operators, remains an area that requires more comprehensive exploration. While robust and practical LfD* solutions for the currently emphasised capabilities continue to promise valuable influence on the assembly industry, the extension to tasks predominantly performed manually can have an exceptional impact on industrial practice. The outlined limitation is particularly evident in methodologies leveraging machine learning techniques. Despite their promising capabilities, only limited knowledge can be gained about their transferability to practical applications.

Given the extensive history of research in this area and the considerable benefits promised by industrial deployment, there is a pressing need to prioritise evaluations using industry-relevant tasks within realistic environments.

**Task Complexity and Diversity –** The use of LfD*-equipped robotic systems, as opposed to specialised machinery designed for defined assembly operations, lies in their enhanced user-friendliness, reduced acquisition costs and, in particular, increased applicability. The latter requires a high degree of adaptability to different task complexities and varieties, which must be anchored in the applied LfD* solution.

Contrary to this industrial requirement, the literature review discloses that the developed solutions predominantly focus on individual subskills with a strong bias toward peg insertion. While the latter indeed covers a substantial portion of industrially performed tasks, there exists a significant shortcoming in the development of LfD* methods that address other skills, particularly joining skills, which are necessary for the completion of many assembly tasks. Studies investigating pick-and-place-related tasks offer conceptual ideas for handling compound tasks, including automatic extraction of abstract task interpretation and appropriate sequencing of subskills. However, proposed frameworks possess mostly limited subskill repertoires which were designed for specialised assembly situations.

In order to address the needs of the industrial sector, it is crucial to broaden the capability scope of LfD* methods to have a higher degree of achievable task complexity as well as a broader range of subskills. Existing industrial taxonomies (Deutsches Institut für Normung, 2003; Maynard et al., 1948; Shneier et al., 2015; Verband Deutscher Ingenieure, 1990) may provide guidance on potentially required skills for compound assembly tasks.

**Generalisability –** Besides the necessary adaptability to distinct task complexities and varieties, the incorporation of LfD* solutions in the industrial sector holds, in particular, the promise of a significant degree of adaptability to diverse task and environmental conditions. The expected generalisability is seen as one of the key enablers of LfD* frameworks to manage the predicted shift from mass-production to mass-customisation.

The analysis of the generalisability of state-of-the-art LfD* solutions shows that recent studies focus on experimental evaluations on encountering environmental discrepancies between demonstrated and executed tasks. This includes, in particular, the spatial scaling capabilities that are intrinsically given in most applied techniques. The reported performances are perceived as appropriate in the context of current industrial requirements. Furthermore, the qualitative analysis shows an emerging cluster dealing with realistic environment dynamics, especially in pick-and-place-related contexts. The learning of corresponding abstract subskill sequences contributes decisively to generalisation capabilities favouring the seamless transfer to realistic industrial tasks.

Nevertheless, these features represent only a fraction of the cognitive abilities that human operators possess. The transferability to like-minded tasks, be it in terms of similar component groups or sequence structures, is an essential property that increases the applicability of LfD* solutions in the field of industrial assembly. This is particularly true with regard to the ability to handle product variants, which is one of the decisive challenges in the automation of industrial assembly. To address this pivotal feature, further research is to be undertaken in the area of cross-product generalisation skills. Although similar objects have been revealed as an emerging domain in the reviewed literature, the analysed studies show the pressing need to extend the investigations toward generalising to similar objects. Machine learning-based approaches promise significant progression in this area, especially through the rapid development of artificial intelligence techniques, but show limited evaluation in practical applications.

**Performance Evaluation –** The primary objectives of industrial assembly lie in its pursuit of efficiency, productivity, and cost-effectiveness (Nof et al., 1997). An established method to analyse the economic benefits of automation solutions is given by the Overall Equipment Effectiveness (OEE) that incorporates productivity losses caused by setup and waiting times, reduced execution speeds, scrap and rework (Roth & Zur Steege, 2015). Alternative performance metrics are provided by standardisation organisations such as the National Institute of Standards and Technology (Shneier et al., 2015).

The conducted literature review revealed the utilisation of different performance metrics to assess proposed LfD* solutions. These include, in particular, quantifying success rates and mastered tolerances in the case of peg insertion tasks. Furthermore, common practice is to compare against competitive LfD* approaches assessing the methods' effectiveness. Only a small part deals with the systems' efficiency, which was evaluated based on processing or execution time. To promote the viability of LfD* solutions for industrial applications, a more comprehensive performance evaluation of the system as a whole is necessary.

In light of the OEE method, the performance evaluation should be extended to the consortium of industry-relevant indicators. As inspiration, the execution duration may be calculated as a ratio of LfD*-trained robot movement speed to its maximal technical speed limit or in comparison to human operators (Maynard et al., 1948). The setup and waiting times may be mapped to the time required to (re-)teach an LfD*-system in relation to its scheduled utilisation time. An optional performance indicator is also seen within the cost-efficiency by conducting a cost-benefit analysis.

In addition, the use of benchmark models provides a convenient way of validating developed capabilities and should therefore be exploited more. Existing benchmark models for assembly tasks are given by the Cranfield set (Collins et al., 1985), the NIST assembly task boards (NIST, 2018) with suggested performance metrics by (Kimble et al., 2020, 2022), the peg insertion setup by (Van Wyk et al., 2018), the metrics for force control by (Falco et al., 2016) and other general manipulation benchmarks and challenges (Calli et al., 2015; Watanabe et al., 2017). If interested parties from the industry are looking for specific skills, the proposal of customised benchmark models is a suggested approach.

**Integration Concepts –** Since their invention, collaborative robots have been an emerging technology that is steadily migrating into the industrial world to assist humans with their work tasks. For a seamless integration of such systems, the safety of humans is inevitable. Apart from physical hazards, are ethical, social and phycological aspects equally important.

In addition to the obstacles above, framing factors outside the technical solutions will determine the ability to transfer the research results into industrial environments. The conducted literature review discloses a dominant focus on understanding the sensorimotor skills to succeed in physical tasks. However, for systematic integration into the working environment, it is equally important to develop a comprehensive framework that considers aspects beyond the technical succession. Accordingly, further research potentials are emerging in the context of integration concepts. This should include the development of intuitive but comprehensive guidelines to familiarise and interact with the robot system reducing potential frustration by taking ethical, social, and phycological aspects into account. Building on multiple in parallel applied communication channels, e.g. extending common approaches with verbal commands, may support the instructors' comfort by improving communication and knowledge transfer. Furthermore, the minimisation of the required demonstration repetition is essential to increase efficiency and reduce the potential for frustration experienced by the instructor.

### 4.3 Limitations of the Systematic Literature Review

With the intention of comparing state-of-the-art achievements of learning from demonstration research to industrial requirements, the systematic literature focuses deliberately on end-to-end solutions performed physically by humans and robots. Consequently, potential impactful theoretical findings may be excluded either intentionally by the authors or unintentionally due to uncontrollable circumstances. One acute reason for unintentional limitation was the COVID-19 pandemic, impacting work physically in laboratories and workplaces between 2020 and 2022. Such exclusions are covered by the criteria EX6 and EX7 (assessment performed in order: EX6 before EX7). Figure 8 illustrates the chronological distribution of excluded studies as a percentage of included+EX6/7. The outlier in 2015 and 2019 are attributable to a small number of two included versus two excluded records. The remaining outlier in 2018 and 2020 are based on two out of seven excluded records. Hence, the COVID-19 pandemic has statistically no exclusive impact on the systematic literature review results. Regarding the deliberate omission of physical experiments, impactful fundamental theoretical achievements without physical evaluation are assumed to be covered by reviews and surveys that emphasise learning from demonstration methodologies (see Table 1). Finally, the chosen *citation requirement* might have excluded valuable studies. However, to the best of the authors' knowledge, the systematic literature review reflects comprehensively current trends and procedures in academic research of learning from demonstration for assembly-related tasks.

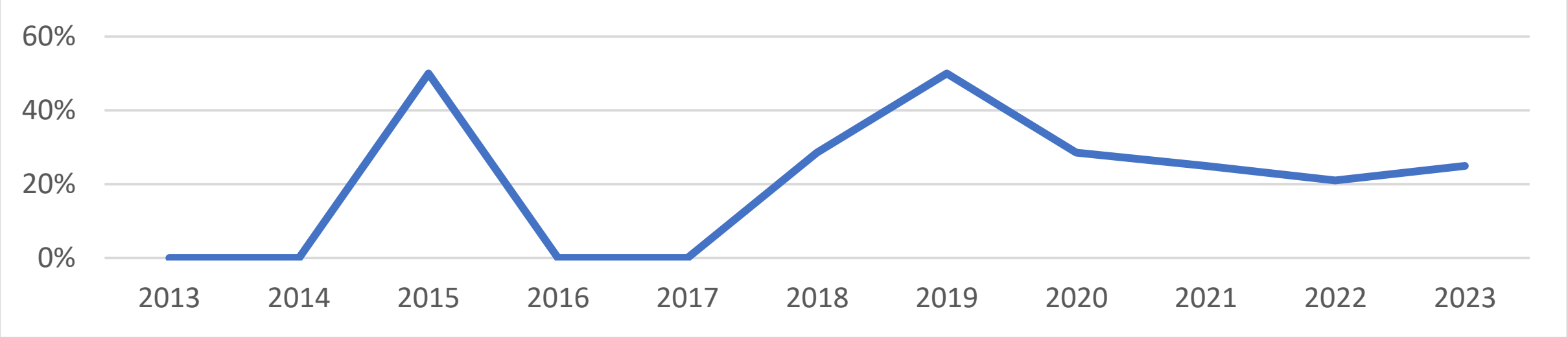

*Figure 8: Chronological distribution of excluded studies based on non-human demonstration (EX6) or non-physical robotic execution (EX7) as a percentage of included+EX6/7 studies*

## 5. Conclusion

Emerging challenges in the assembly industry promote the implementation of assistive robotic solutions that encounter growing product demand and variability. A promising method lies in the simplification and improvement of robot programming in terms of accessibility and generalisation capabilities, which reduces the need for robot experts and significantly enhances applicability in a broader range of assembly applications. These characteristics can be attributed to the concept of learning from demonstration and its variations, which despite consistent attention in research, has not yet gained a significant foothold in the repetitive assembly industry.

The present systematic literature review aimed to explore the factors contributing to the limited adoption of learning from demonstration solutions within the repetitive assembly industry. Conducted physical experiments were quantitatively and qualitatively analysed to determine current trends, interests and achievements within relevant contributions of the last decade. A comparison was drawn to the industry-established four-step instruction technique for human operators to identify synergies and discrepancies of LfD* solutions with industrial assembly practice. Promising synergetic concepts were found in demonstration methods, approaches for the interactive elaboration of the task's understanding and the subsequent improvement during independent execution. However, major obstacles were identified primarily in real-world evaluations of practicability and task complexity. While developed LfD* solutions provide sufficient generalisation in terms of environmental constraints, the emerging field of generalising over product variations requires more attention to encounter mass customisation. An additional obstacle is expected beyond the technical development in integrational aspects.

Overall, LfD* experimental evaluations have reached a significant level of maturity in which the conceptual understanding and exploration of capabilities have been sufficiently proven to be reasonable for encountering challenges in several practical domains. The necessary next step to enable the value of LfD* solutions for industrial practices requires their profound deployments in real-world applications.

## 6. Acknowledgements

The research leading to these results was supported by the University of Technology Sydney's President's Scholarship, International Research Scholarship and the Industrial Transformation Training Centre (ITTC) for Collaborative Robotics in Advanced Manufacturing (also known as the Australian Cobotics Centre) funded by ARC (Project ID: IC200100001).

## 7. Appendix

| *Reference, Year* | *Avg. Citation Count* | *Demonstration Method* | *Demonstration Quantity* | *Learning Method* | *Assembly Skill* | *Practicability* | *Application Scenario* | *Generalisability* | *Performance Measure* |
|---|---|---|---|---|---|---|---|---|---|
| (Yan et al., 2023) | -- | Kinesthetic Teaching | 10 | Costs/Rewards - Trajectory Optimisation | peg insertion (0.2 mm tolerance) | unrelated | --- | spatial scaling<br>task uncertainties | success rate (100% / 10)*<br>efficiency<br>effectiveness |
| (Ahn et al., 2023) | -- | Kinesthetic Teaching | unspecified | Costs/Rewards - Trajectory Optimisation | peg insertion (0.1 mm tolerance) | unrelated | --- | spatial scaling<br>similar object | success rate (96% / 100 same object, 95% / 100 similar object)*<br>effectiveness |
| (Eiband et al., 2023) | -- | Kinesthetic Teaching | 1 | Policy - Trajectory | gluing<br>pick-and-place | unrelated | --- | unspecified | success rate (83% / 6)*<br>effectiveness |
| (Aljaz Kramberger et al., 2022) | 6.0 | Kinesthetic Teaching | "several" | Plan - Primitive Sequence | interlocking | related | timber structure assembly | spatial scaling<br>task uncertainties | success rate (93% / 100)* |
| (Hu et al., 2022) | 3.0 | Kinesthetic Teaching | "multiple" | Policy - Trajectory | peg insertion (0.5 mm tolerance) | practical | PCB assembly | spatial scaling | success rate (100% / 9)*<br>efficiency<br>effectiveness |
| (Davchev et al., 2022) | 3.0 | Teleoperation | single | Policy - Trajectory | peg insertion (0.4 mm tolerance) | related | RJ-45 connector | spatial scaling<br>temporal scaling<br>similar object | success rate (86.9% / -)*<br>efficiency<br>effectiveness |
| (Y. Zhang et al., 2022) | 3.0 | Kinesthetic Teaching | 4 | Policy - Trajectory | pick-and-place | unrelated | --- | spatial scaling | success |
| (Caldarelli et al., 2022) | 3.0 | Kinesthetic Teaching | "few" | Policy - Trajectory | peg insertion (tolerance not specified) | unrelated | --- | unspecified | success |
| (Huang et al., 2022) | 2.0 | Passive Observation | "multiple" | Policy - Trajectory | sewing | related | personalised stent grafts | spatial scaling<br>task uncertainties<br>path optimisation | success |
| | | | | | | | | similar objects | |
| (Yan Wang et al., 2022) | 2.0 | Kinesthetic Teaching | 5 | Policy - Trajectory | peg insertion (1 mm tolerance) | related | USB stick and power plug | spatial scaling<br><br>similar objects | success rate (90% / 20)*<br><br>efficiency |
| (Ti et al., 2022) | 2.0 | Passive Observation | "multiple" | Policy - Trajectory | peg insertion (0.3 mm tolerance) | unrelated | --- | spatial scaling | success rate (93.8% / 80)*<br><br>effectiveness |
| (Ma et al., 2022) | 1.0 | Teleoperation | 8 | Costs/Rewards - Trajectory Optimisation | peg insertion (0.01 mm tolerance, interference fit) | unrelated | --- | similar object | success rate (100% / 20)*<br><br>effectiveness |
| (Shetty et al., 2022) | 1.0 | Kinesthetic Teaching | "set" | Policy - Low-level Actions | peg insertion (0.89 mm tolerance) | unrelated | --- | spatial scaling<br><br>task uncertainties | success rate (100% / 20)*<br><br>efficiency |
| (Su et al., 2022) | 1.0 | Teleoperation | 10 | Policy - Trajectory | peg insertion (0.1 mm tolerances) | unrelated | --- | spatial scaling | success<br><br>effectiveness |
| (Meszaros et al., 2022) | 1.0 | Teleoperation + Kinesthetic Teaching | single | Policy - Trajectory | pick-and-place | unrelated | --- | spatial scaling<br><br>similar object | success rate(82% / 50)*<br><br>efficiency |
| (Deng et al., 2022) | 0.0 | Passive Observation | 15 | Policy - Trajectory | pick-and-place | unrelated | --- | unspecified | success rate (90% / 10)*<br><br>accuracy error of trajectory |
| (Hongmin Wu et al., 2022) | 0.0 | Kinesthetic Teaching | 4 | Policy - Trajectory | pick-and-place | unrelated | --- | spatial scaling | success rate (85% / 20)*<br><br>efficiency<br><br>effectiveness |
| (Stepputtis et al., 2022) | 0.0 | Teleoperation | 30 | Policy - Trajectory | peg insertion (6 mm tolerance) | unrelated | --- | spatial scaling<br><br>task uncertainties | success rate (90% / 30)*<br><br>effectiveness |
| (Pellois & Brüls, 2022) | 0.0 | Passive Observation | single | Policy - Trajectory | pick-and-place | unrelated | --- | path optimisation | success rate (100% / 10)* |
| (Pinosky et al., 2022) | 0.0 | Teleoperation | 3 | Costs/Rewards - Trajectory Optimisation | stacking | unrelated | --- | spatial scaling | success<br><br>effectiveness |
| (Kang & Oh, 2022) | 0.0 | Passive Observation | 30 | Plan - Primitive Sequence | stacking | unrelated | --- | unspecified | success rate (100% / 5)* |
| (Guo & Burger, 2022) | 0.0 | Kinesthetic Teaching | 6 | Plan - Primitive Hierarchy | pick-and-place<br><br>bin-sorting | unrelated | --- | spatial scaling<br><br>sequence optimisation | success rate (100% / 20)*<br><br>effectiveness |
| (Keipour et al., 2022) | 0.0 | Passive Observation | single | Plan - Primitive Sequence | wiring | related | NIST Assembly Board #3 | spatial scaling | success<br><br>efficiency |
| (Meattini et al., 2022) | 0.0 | Kinesthetic Teaching | single | Policy - Trajectory | wiring | unrelated | --- | task uncertainties | success rate (100% / 10)* |
| (Arguz et al., 2022) | 0.0 | Kinesthetic Teaching | single | Policy - Trajectory | peg insertion (1.8 mm tolerance) | unrelated | --- | spatial scaling | success rate (62.5% / -)* |
| (Jha et al., 2022) | 0.0 | Teleoperation | single | Policy - Trajectory | peg insertion (2 mm tolerance) | unrelated | --- | spatial scaling<br><br>task uncertainties | success rate (96% / 50 moving hole, 98.6% / 80 error added at known location)*<br><br>effectiveness |
| (Iovino et al., 2022) | 0.0 | Kinesthetic Teaching | 3 | Plan - Primitive Sequence | pick-and-place | unrelated | --- | spatial scaling | success rate (75% / 4)* |
| (Yu & Chang, 2022) | 0.0 | Kinesthetic Teaching | unspecified | Costs/Rewards - Trajectory Optimisation | screwing<br><br>stacking | unrelated | --- | spatial scaling<br><br>path adjustment | success |
| (W. Wang et al., 2022) | 0.0 | Kinesthetic Teaching | 6 | Policy - Trajectory | peg insertion (tolerance not specified) | unrelated | --- | spatial scaling | success<br><br>effectiveness |
| (Xu et al., 2022) | 0.0 | Passive Observation | unspecified | Costs/Rewards - Trajectory Optimisation | stacking | unrelated | --- | spatial scaling | success rate (96% / -)*<br><br>effectiveness |
| (Y. Chen et al., 2022) | 0.0 | Passive Observation | 3 | Plan - Primitive Sequence | stacking | unrelated | --- | sequence optimisation | success rate (77% / -)*<br><br>efficiency |
| (D. Liu et al., 2022) | 0.0 | Teleoperation | 15 | Plan - Primitive Sequence | stacking<br><br>pick-and-place | unrelated | --- | spatial scaling | success rate (100% / 7 stacking, 100% / 7 |
| | | | | | | | | | pick-and-place)*<br><br>efficiency<br><br>effectiveness |
| (Hongtao Wu et al., 2022) | 0.0 | Passive Observation | 10 | Plan - Primitive Sequence | stacking | unrelated | --- | spatial scaling | success rate (> 90% / 10 trained tasks, >50% / 10 unseen tasks)*<br><br>effectiveness |
| (X. Zhang et al., 2021) | 10.0 | Kinesthetic Teaching | 30 | Costs/Rewards - Inverse Reinforcement Learning | pick-and-place | unrelated | --- | spatial scaling | success<br><br>accuracy |
| (Yan Wang et al., 2021a) | 4.5 | Teleoperation | 8 | Policy - Trajectory | peg insertion (1 mm tolerance)<br><br>peg insertion (unspecified) | practical<br><br>related | condenser assembly<br><br>HDMI insertion | task uncertainties | success rate (100% / 20)*<br><br>efficiency<br><br>effectiveness |
| (Ji et al., 2021) | 4.5 | Kinesthetic Teaching + Passive Observation | unspecified | Plan - Primitive Sequence | peg insertion (0.3mm tolerances)<br><br>peg insertion (0.01 mm tolerance)<br><br>bin-picking<br><br>bolting | practical | power breaker assembly + set-top box assembly | spatial scaling<br><br>path optimisation | success rate: (98% / 100 grasping, 97% / 100 inserting)* |
| (Hu et al., 2021) | 4.0 | Passive Observation | 9 | Policy - Trajectory | peg insertion (0.42 mm tolerance) | practical | PCB assembly | spatial scaling | success |
| (Y. Q. Wang et al., 2021) | 3.5 | Kinesthetic Teaching | 5 | Policy - Trajectory | pick-and-place | unrelated | --- | spatial scaling<br><br>path adjustment | success<br><br>efficiency<br><br>effectiveness |
| (Yan Wang et al., 2021b) | 3.0 | Teleoperation | 8 | Policy - Trajectory | peg insertion (1 mm tolerance)<br><br>peg insertion (unspecified) | practical<br><br>related | condenser assembly<br><br>HDMI connector | task uncertainties | success |
| (Berscheid et al., 2020) | 7.3 | Passive Observation | single | Costs/Rewards - Trajectory Optimisation | pick-and-place<br><br>peg insertion (1mm tolerance)<br><br>bin-picking | unrelated | --- | spatial scaling<br><br>path adjustment<br><br>similar objects | success rate (95% / - grasping rate, 72% / 10 peg insertion, 86% / - select trained objects)*<br><br>accuracy |
| (Ugur & Girgin, 2020) | 6.3 | Kinesthetic Teaching | 9 | Policy - Trajectory | pick-and-place | unrelated | --- | path adjustment | success |
| (Cho et al., 2020) | 5.0 | Kinesthetic Teaching | 8 | Policy - Trajectory | peg insertion (0.2mm tolerance) | unrelated | --- | spatial scaling<br><br>task uncertainties<br><br>similar objects | success<br><br>efficiency |
| (N. Liu et al., 2020) | 4.0 | Kinesthetic Teaching | 5 | Policy - Trajectory | peg insertion (20mm H7f7 tolerance) | unrelated | --- | spatial scaling<br><br>task uncertainties | success rate (100% / 3)* |
| (Gubbi et al., 2020) | 3.3 | Teleoperation | 8 | Policy - Low-level Actions | peg insertion (0.006 mm tolerance) | unrelated | --- | task uncertainties | success |
| (Duque et al., 2019) | 14.3 | Passive Observation | 10 | Policy - Trajectory | peg insertion (tolerance not specified) | unrelated | --- | spatial scaling | success rate (86.7% / 30)* |
| (Qin et al., 2019) | 7.0 | Teleoperation | 4 to 7 | Plan - Primitive Sequence | peg insertion (0.01 mm clearance fit)<br><br>gluing<br><br>peg insertion (0.05 mm clearance fit) | practical | sleeve-cavity and coil-cylinder assembly | spatial scaling<br><br>task uncertainties | success<br><br>accuracy<br><br>effectiveness |
| (Savarimuthu et al., 2018) | 8.2 | Passive Observation + Teleoperation | single | Plan - Primitive Hierarchy | peg insertion (tolerance not specified) | related | Cranfield benchmark assembly | spatial scaling<br><br>task uncertainties | success rate (50% / -)* |
| (Ghalamzan E. & | 6.4 | Kinesthetic Teaching | 2 | Policy - Trajectory | pick-and-place | unrelated | --- | spatial scaling | success rate (100% / 5)* |
| Ragaglia, 2018) | | | | | | | | path adjustment | |
| (Gu et al., 2018) | 4.4 | Kinesthetic Teaching + Passive Observation | "multiple" | Plan - Primitive Sequence | hammering<br><br>bolting<br><br>screwing | unrelated | --- | spatial scaling | success rate (90% / 20 hammering, 60% / 20 screwing, 75% / 20 bolting)*<br><br>effectiveness |
| (Gašpar et al., 2018) | 4.4 | Passive Observation | 99 | Policy - Trajectory | peg insertion (tolerance not specified) | unrelated | --- | spatial scaling<br><br>temporal scaling | success<br><br>accuracy |
| (Yue Wang et al., 2018) | 3.4 | Passive Observation | single | Plan - Primitive Sequence | pick-and-place<br><br>screwing | practical | Switch assembly | spatial scaling | success rate (100% / 4 pick-and-place, 75% / 4 screwing)* |
| (Perez-D'Arpino & Shah, 2017) | 7.5 | Teleoperation | single | Plan - Primitive Sequence | pick-and-place | unrelated | --- | unspecified | success rate (100% / 10)* |
| (Wan et al., 2017) | 6.2 | Passive Observation | single | Policy - Trajectory | peg insertion (0.04 mm tolerance) | unrelated | --- | similar object | Success |
| (Aljaž Kramberger et al., 2017) | 5.8 | Kinesthetic Teaching | 100 | Policy - Trajectory | peg insertion (1 mm tolerance) | unrelated | --- | spatial spacing<br><br>task uncertainties | success<br><br>efficiency<br><br>effectiveness |
| (Pervez et al., 2017) | 5.5 | Teleoperation | 4 | Policy - Trajectory | peg insertion (tolerance not specified) | unrelated | --- | spatial scaling | success rate (100% / 4)* |
| (Sefidgar et al., 2017) | 4.8 | Passive Observation | unspecified | Plan - Primitive Sequence | pick-and-place | unrelated | --- | spatial scaling | success rate (90% / 20)/* |
| (Haage et al., 2017) | 3.2 | Passive Observation | single | Plan - Primitive Sequence | pick-and-place | practical | PCB assembly | spatial scaling | success<br><br>effectiveness |
| (Tang et al., 2016) | 6.7 | Passive Observation | 50 | Policy - Trajectory | peg insertion (25mm H7h7 tolerance + 1mm chamfer) | unrelated | --- | spatial scaling<br><br>task uncertainties | success rate (98% / 25)* |
| (Niekum et al., 2015) | 16.3 | Kinesthetic Teaching | 8 | Policy - Trajectory | screwing | unrelated | --- | spatial scaling | success rate (90% / 10)* |
| (Abu-Dakka et al., 2015) | 10.9 | Teleoperation + Kinesthetic Teaching | single | Policy - Trajectory | peg insertion (tolerance not specified) | related | Cranfield benchmark assembly | spatial scaling<br><br>task uncertainties | success rate (86% / 50)*<br><br>efficiency |
| (Abu-Dakka et al., 2014) | 5.1 | Kinesthetic Teaching | single | Policy - Trajectory | peg insertion (tolerance not specified) | related | Cranfield benchmark assembly | spatial scaling<br><br>task uncertainties | success rate (100% / 50 shaft and round pegs, 96% / 50 square pegs)* |

*Appendix A: Comprehensive Overview of included References (* success rate / attempts)*